\pdfoutput=1
\documentclass[11pt]{article}
\usepackage[final]{acl}
\usepackage{times}
\usepackage{latexsym}
\usepackage[T1]{fontenc}
\usepackage[utf8]{inputenc}
\usepackage{microtype}
\usepackage{graphicx}
\usepackage{booktabs}
\usepackage{comment}

\usepackage{amsmath}
\usepackage{amssymb}
\usepackage{todonotes}
\usepackage{xspace}
\usepackage{multirow}
\usepackage{enumitem}
\usepackage{tabularx}
\usepackage{float}

\makeatletter
\g@addto@macro\normalsize{%
    \setlength{\abovedisplayskip}{-6pt}   
    \setlength{\abovedisplayshortskip}{-7pt}  
    \setlength{\belowdisplayshortskip}{-0.5pt}  
}
\makeatother
\newcommand{\modelname}{\textit{DAM}\xspace}
\newcommand{\maxlength}{\textit{PCL}\xspace}

\title{\modelname: Dynamic Attention Mask for Long-Context Large Language Model Inference Acceleration}

\author{
  Hanzhi Zhang\textsuperscript{1},
  Heng Fan\textsuperscript{1},
  Kewei Sha\textsuperscript{2},
  Yan Huang\textsuperscript{1},
  Yunhe Feng\textsuperscript{1} \\[1ex]
  \textsuperscript{1}LLaVi Lab, Department of Computer Science \& Engineering; 
  \textsuperscript{2}Department of Data Science \\ 
  University of North Texas \\[1ex]
  \texttt{\{hanzhi.zhang,\,heng.fan,\,kewei.sha,\,yan.huang,\,yunhe.feng\}@unt.edu}
}

\begin{document}

\maketitle

\begin{abstract}
Long-context understanding is crucial for many NLP applications, yet transformers struggle with efficiency due to the quadratic complexity of self-attention. Sparse attention methods alleviate this cost but often impose static, predefined masks, failing to capture heterogeneous attention patterns. This results in suboptimal token interactions, limiting adaptability and retrieval accuracy in long-sequence tasks. 
This work introduces a dynamic sparse attention mechanism that assigns adaptive masks at the attention-map level, preserving heterogeneous patterns across layers and heads. Unlike existing approaches, our method eliminates the need for fine-tuning and predefined mask structures while maintaining computational efficiency. By learning context-aware attention structures, it achieves high alignment with full-attention models, ensuring minimal performance degradation while reducing memory and compute overhead.
This approach provides a scalable alternative to full attention, enabling the practical deployment of large-scale Large Language Models (LLMs) without sacrificing retrieval performance.
\modelname is available at: \textcolor{blue}{\url{https://github.com/HanzhiZhang-Ulrica/DAM}}. 
\end{abstract}

\section{Introduction}

\begin{figure*}[htbp]
    \centering
    \vspace{-0.8cm}
    \includegraphics[width=\textwidth]{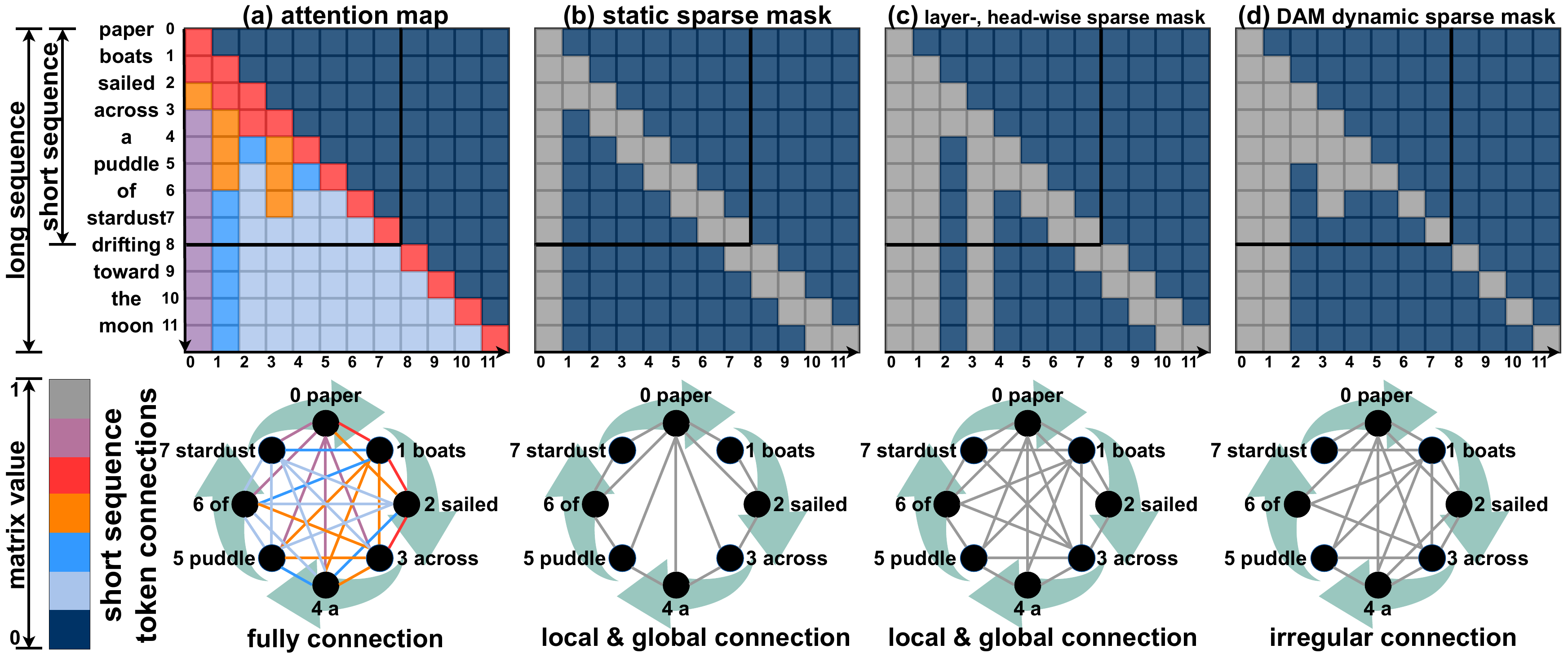}
    \vspace{-0.6cm}
    \caption{Attention patterns from $queries \times keys$. The short input sequence "paper boats sailed across a puddle of stardust" is a subset of the longer input sequence "paper boats sailed across a puddle of stardust drifting toward the moon". (a) Each query attends to all keys, and longer attention patterns are extensions of short patterns. (b) Static attention map captures same classical global and sliding-window patterns to attention maps from every layers and heads. (c) Different masks are assigned to corresponding maps with predefined patterns. (d) Heterogeneous map remains feature patterns from each attention map.}
    \label{fig:task}
    \vspace{-0.4cm}
\end{figure*}

Understanding long contexts is essential for document summarization, question answering, and retrieval-augmented generation. Long-context NLP applications power legal analysis, financial reporting, and knowledge graph construction, where maintaining coherence across tokens is critical. However, existing LLMs struggle with long sequences due to inefficiency in self-attention.

LLMs leverage the Transformer architecture, which models long-range token dependencies through self-attention, enabling direct interactions between all tokens. Multi-head attention enhances expressivity by capturing multiple token relationships, while positional embeddings preserve order information. Dynamic token representations allow contextual meaning to evolve across layers, ensuring coherent and accurate long-term recall.

However, transformers process long contexts inefficiently. Quadratic complexity makes full attention computationally prohibitive, forcing models to truncate inputs, leading to information loss in long-document tasks. Attempts to mitigate this through fixed context windows or uniform attention fail to distinguish between critical and redundant information. Meanwhile, streaming applications suffer from recomputing at every step, as each newly generated token attends to all previous tokens. This results in redundant computation, growing memory usage, and increased latency, making real-time processing impractical.

The inability to process long contexts also directly impacts businesses and researchers. Legal and financial institutions rely on AI to analyze contracts and reports\cite{pingiliai}, but truncated inputs cause critical details to be lost. AI assistants in customer service forgetting past interactions may fail to maintain coherent conversations. Researchers pushing transformer efficiency face skyrocketing computational costs, making large-scale deployment unsustainable. Without an efficient solution, enterprises must rely on costly and ineffective workarounds like document chunking, which may destroy contextual coherence.

Sparse attention is an efficiency-driven approach to mitigating long-context inefficiency in generative LLMs by enforcing structured sparsity. Figure~\ref{fig:task}(b) illustrates static sparse attention methods that reduce computational cost by enforcing fixed-span sliding window and global masks across all heads and input lengths. This approach improves efficiency but sacrifices flexibility, forcing models to rely only on local interactions. As a result, these models fail to adapt to long-range dependencies, reducing accuracy in complex retrieval and reasoning tasks. Figure ~\ref{fig:task}(c) improves flexibility by assigning different masks to layers and heads, removing the need for fine-tuning. However, it assumes attention structures can be predefined, failing to capture heterogeneous token interactions that emerge dynamically. The result is a rigid sparsity pattern that still requires processing all sequence lengths, making it resource-intensive for long-context applications.

This work introduces \modelname{}, a novel framework for dynamic sparse attention, as illustrated in Figure~\ref{fig:task}(d).  \modelname generates adaptive sparse attention masks at the granularity of individual attention maps, thereby capturing both layer-specific structural patterns and input-dependent variations in attention. In contrast to prior approaches that often rely on fixed or globally-defined sparsity patterns, \modelname{} preserves the heterogeneity of attention patterns across different layers and heads, leading to improved expressiveness.  Furthermore, the method eliminates the need for manual, task-specific fine-tuning of the sparsity structure, while maintaining the computational benefits of sparse attention.

Our contributions are summarized as follows:
\begin{itemize}[nosep,left=0pt] 
    \item We propose a dynamic sparse attention framework that assigns distinct, adaptive sparse masks to each attention map, preserving heterogeneous patterns across heads and layers.
    \item Our approach is fine-tuning-free and generalizes seamlessly to varying input lengths, eliminating the need for manual sparsity pattern design.
    \item We incorporate a flexible "true mask" mechanism to focus attention on relevant regions, reducing unnecessary computations on padding tokens or less informative areas. 
    \item We demonstrate that \modelname achieves performance comparable to full-attention models while improving computational efficiency. 
\end{itemize}

%%%%%%%%%%%%%%%%%%%%%%%%%%%%%%%%%%%%%%%%%%%%%%%%%%%
%%%%%%%%%%%%%%%%%%%%%%%%%%%%%%%%%%%%%%%%%%%%%%%%%%%

\section{Related Work}

Attention mechanisms enable transformers to model dependencies across sequences but introduce computational challenges at scale. Researchers have explored multiple strategies to address these inefficiencies, including KV-caching for faster inference, sparse and hierarchical attention for memory reduction, state-space models for efficient streaming, and hybrid architectures for improved long-term memory tracking. While these approaches enhance scalability, each introduces trade-offs that limit their applicability to long-sequence processing.

KV-cache enhances autoregressive decoding by storing key and value representations from previous steps, allowing reuse instead of recomputing attention scores for all tokens~\cite{ge2023model,li2024snapkv,zhang2024h2o,zhao2024buzz,chen2024nacl,liu2024scissorhands,adnan2024keyformer,ge2023model}. This reduces redundant computation and accelerates inference but increases memory usage, limiting scalability for long sequences~\cite{zhang2024unifying,ye2024chunkattention,hu2024memserve}. Cache management adds complexity, and performance gains depend on reuse efficiency~\cite{zheng2024batchllm,zheng2024sglang,xiong2024layerkv,gao2024cost}. While KV-cache mitigates inefficiencies in autoregressive generation, it does not reduce the fundamental complexity of self-attention.

Sparse attention reduces token interactions to improve efficiency~\cite{child2019generating,yun2020n,ho2019axial}. Static sparse attention applies predefined masks across all processed sentences to lower computational cost and improve hardware utilization~\cite{roy2021efficient,kitaev2020reformer,tay2019lightweight,choromanski2020masked}. Common approaches include global, sliding window, and random masks, with local attention patterns enabling KV-cache eviction beyond the attention span to reduce memory usage~\cite{beltagy2020longformer,ainslie2004etc,zaheer2020big}. However, static masks remain uniform across layers and heads, ignoring token-specific dependencies. This rigidity leads to information loss in long-sequence tasks, where retrieval accuracy relies on adapting attention spans dynamically.

Other strategies generate distinct masks by leveraging statistical information, defining role-specific constraints for attention heads, or introducing context-dependent sparsity~\cite{wang2020multiheadselfattentionroleguidedmasks,fusemsa,fu2024moa,correia2019adaptivelysparsetransformers}. While these approaches increase flexibility, they fail to dynamically capture heterogeneous attention within individual maps and still process all sequence lengths, raising resource costs.

To introduce flexibility, some approaches assign different predefined sparse patterns to layers and heads~\cite{fu2024moa,wang2020multiheadselfattentionroleguidedmasks,fusemsa,correia2019adaptivelysparsetransformers}. They select masks based on input length, improving adaptability without requiring fine-tuning. However, it assumes optimal attention structures are predefined, missing dynamic token interactions, and require evaluating multiple sequence lengths, thereby increasing computational overhead.
This limitation motivates methods to infer sparse structures without exhaustive manual design or repeated inference.

%%%%%%%%%%%%%%%%%%%%%%%%%%%%%%%%%%%%%%%%%%%%%%%%%%%%%%%%%%%%%%
%%%%%%%%%%%%%%%%%%%%%%%%%%%%%%%%%%%%%%%%%%%%%%%%%%%%%%%%%%%%%%
\section{Preliminaries}

Transformer models adopt the scaled dot-product attention mechanism, a core component for capturing relationships between tokens in a sequence \cite{vaswani2017attention}. Attention scores are calculated as $S = \frac{QK^\top}{\sqrt{d_k}}$, where \( Q \in \mathbb{R}^{n \times d_k} \) and \( K \in \mathbb{R}^{m \times d_k} \) denote the query and key matrices, respectively. Here, \( n \) and \( m \) denote the number of query and key/value vectors, while \( d_k \) represents the dimensionality of each key/query vector. The resulting matrix \( S \in \mathbb{R}^{n \times m} \) contains the unnormalized attention logits, representing the pairwise similarities between queries and keys. The scaling factor \( \frac{1}{\sqrt{d_k}} \) is crucial for maintaining numerical stability during training, preventing the dot products from growing excessively large, which can lead to vanishing gradients during backpropagation. This scaling mitigates issues caused by large variances in the logits, particularly when applying a masking operation.

The computation of attention scores for all pairs of tokens has a quadratic time complexity of \( \mathcal{O}(n^2) \) with respect to the sequence length, which becomes computationally expensive for long sequences. Sparse attention mechanisms address this computational bottleneck by imposing structured sparsity on the attention matrix. This is achieved through a binary mask \( M_{\ell, h} \in \{0, 1\}^{n \times m} \) for each layer \(\ell\) and head \(h\), defined as:

\begin{equation*}
M_{\ell, h, i, j} =
\begin{cases}
1, & \text{if token } i \text{ attends to token } j \\ 
   & \text{in layer } \ell \text{ and head } h, \\
0, & \text{otherwise}.
\end{cases}
\end{equation*}

\vspace{-0.2cm}

The mask \( M_{\ell, h} \) is applied element-wise to the attention logits $S' = S \odot M_{\ell, h}$, where \( \odot \) denotes the Hadamard product (element-wise multiplication). This effectively prevents attention between specific token pairs. The masked attention logits are then normalized using the softmax function:

\begin{equation*}
A_{ij} = \frac{\exp(S'_{ij})}{\sum_{k=1}^{m} \exp(S'_{ik})}.
\end{equation*}

\vspace{+0.2cm}

The output of the attention mechanism is computed as a weighted sum of the values, where \( V \in \mathbb{R}^{m \times d_v} \) is the value matrix as $O = A V$.

While sparse attention mechanisms substantially improve computational efficiency, they inherently restrict the model's ability to learn long-range dependencies by limiting token interactions. A key limitation of many sparse attention approaches is their reliance on \textit{fixed} sparsity patterns. Such patterns are unable to adapt to the dynamic nature of attention, including variations in sequence length and the diversity of attention distributions across different inputs. This rigidity can result in a significant reduction in performance, especially when dealing with long sequences or tasks requiring the modeling of intricate relationships. Moreover, predefined sparse attention structures like sliding window \cite{beltagy2020longformerlongdocumenttransformer} or global attention \cite{liu2021globalattentionmechanismretain} often overlook the critical variations in attention patterns that occur across different layers and heads within the network. The optimal set of token interactions evolves across layers, rendering fixed sparsity patterns a bottleneck. 
This motivates the need for a dynamic, structure-aware sparsity mechanism that adapts to position-wise attention patterns while maintaining compatibility with pretrained transformer architectures.

%%%%%%%%%%%%%%%%%%%%%%%%%%%%%%%%%%%%%%%%%%%%%%%%%%%%%%%%%%%%%%
%%%%%%%%%%%%%%%%%%%%%%%%%%%%%%%%%%%%%%%%%%%%%%%%%%%%%%%%%%%%%%
\section{Dynamic Attention Masks (\modelname)}

This section introduces our proposed Dynamic Attention Mask (\modelname) mechanism. We first motivate the design by illustrating the dynamic nature of attention patterns across layers and heads in Transformer models. Then, we detail the architecture of \modelname, and finally, we describe its integration into the standard Transformer framework.

\subsection{Dynamic Attention Patterns}
\label{subsec:dynamic_patterns}

\begin{figure}[htbp]
    \vspace{-0.1cm}
    \centering
    \includegraphics[width=\linewidth]{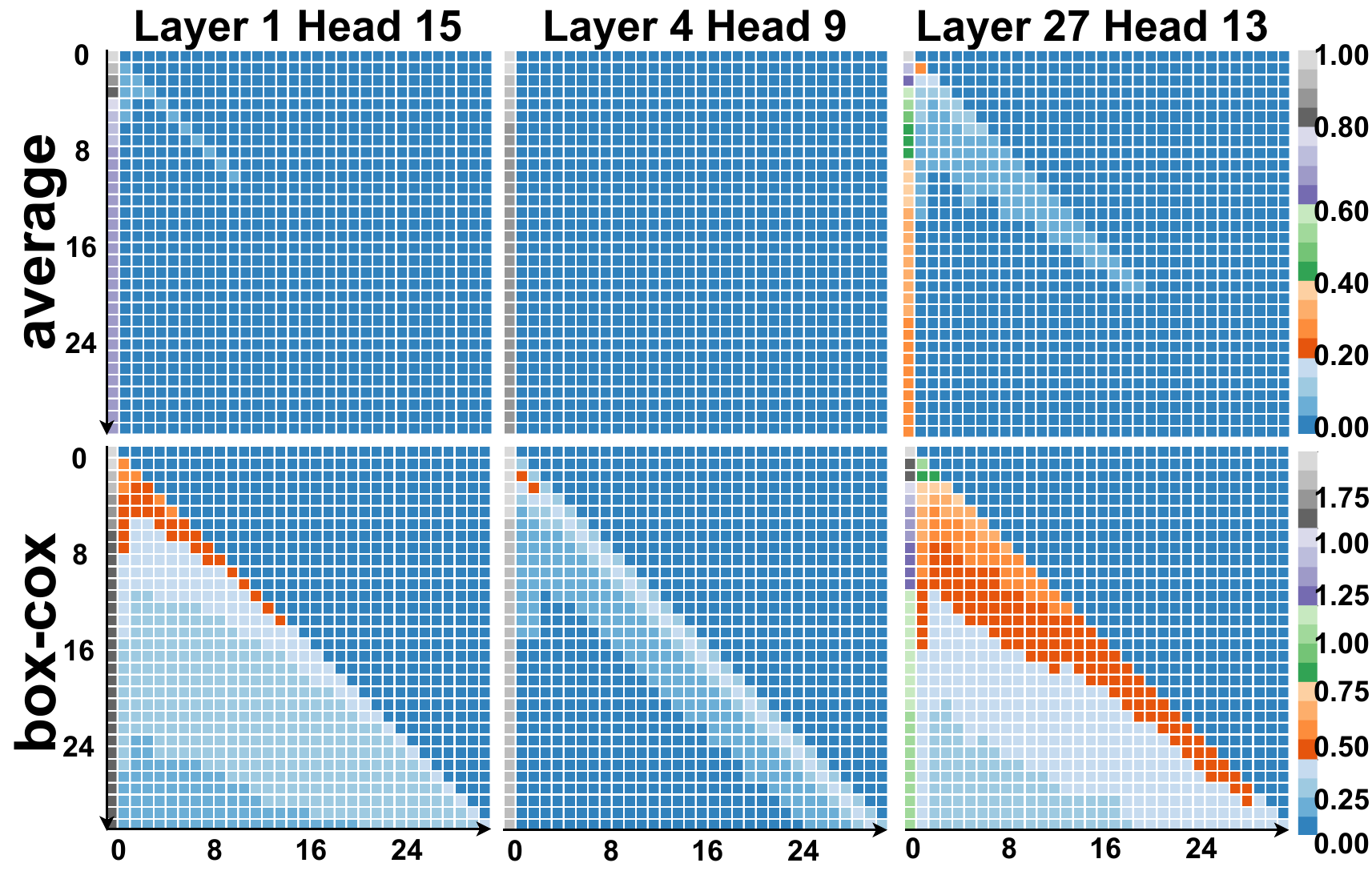}
    \vspace{-0.6cm}
    \caption{Visualization of dynamic attention patterns across layers and heads. The figure compares the effect of averaging (top row) and applying the Box-Cox transformation (bottom row) to attention values from a LLaMA 3.2 3B Instruct model on the Multi-News dataset. The Box-Cox transformation enhances the visibility of dynamic patterns.}
    \label{fig:feature_amplification}
    \vspace{-0.3cm}
\end{figure}

Prior research has investigated and validated the existence of dynamic attention patterns across attention heads and layers in Transformer models \cite{goindani2021dynamicheadimportancecomputation, xiao2024improvingtransformersdynamicallycomposable}. To visualize these patterns, we analyze attention maps obtained from a LLaMA 3.2 3B Instruct model \cite{meta2024llama3.2} evaluated on the Multi-News summarization benchmark \cite{DBLP:journals/corr/abs-1906-01749}.  Figure~\ref{fig:feature_amplification} presents these attention maps, revealing inconsistencies in the underlying sparse structures across different heads and layers. The top row of Figure~\ref{fig:feature_amplification} displays the average attention values across the dataset. While these average maps suggest the presence of dynamic patterns, the patterns themselves are not readily discernible, hindering a deeper understanding and impeding the design of effective sparsity-inducing techniques.

We posit that enhancing the contrast between significant and less significant attention values can reveal these dynamic patterns more clearly. Specifically, we aim to preserve the largest attention values (e.g., those corresponding to the leftmost column in each attention map), while simultaneously accentuating the intermediate values (e.g., those distributed along the diagonal) and differentiating them from the smallest values (e.g., those in the bottom-left regions). To achieve this, we evaluated \textbf{nine different transformation methods} (more details in the Appendix \ref{appx:trans_methods}) and found the Box-Cox transformation \cite{box1964analysis} consistently yielded the most informative visualizations, as shown in the bottom row of Figure~\ref{fig:feature_amplification}.

The Box-Cox transformation enhances the visualization clarity of attention maps and \textbf{amplifies small and medium attention values}, making subtle yet important structural patterns more discernible, while \textbf{preserving the scale of larger values without introducing distortion}. It directly facilitates the intuitive selection and tuning of the threshold parameters (\(\tau\) in Section \ref{sec:true_mask} and \(\mu\) in Section \ref{sec:mask_generation}).

\subsection{Two-Stage Dynamic Attention Masks}

\begin{figure*}[htbp]
    \centering
    \vspace{-0.3cm}
    \includegraphics[width=\textwidth]{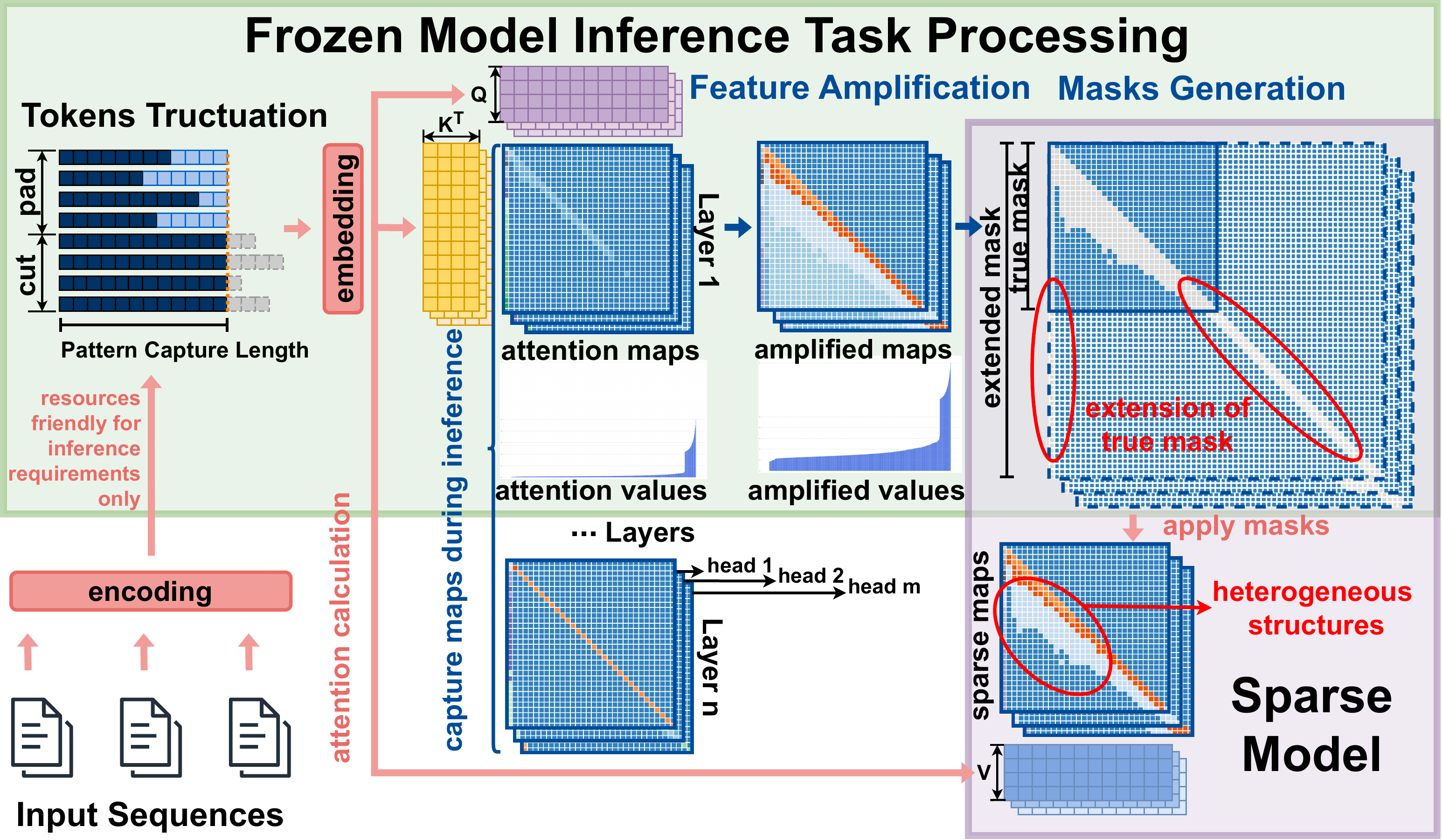}
    \vspace{-0.6cm}
    \caption{Two-stage \modelname overview. The first stage extracts full attention patterns from sequences of varying lengths, applies a Box-Cox transformation, and generates masks that capture essential dependencies. The second stage applies these masks to a sparse model, enabling efficient inference while preserving key attention structures.}
    \label{fig:framework}
    \vspace{-0.4cm}
\end{figure*}

\modelname enhances the efficiency of Transformer models by learning adaptive sparse attention masks. It addresses the limitation of predefined sparsity patterns, which can discard valuable, low-magnitude connections, by dynamically adjusting the attention mask based on observed attention patterns. The framework operates in two stages in Figure \ref{fig:framework}. First, a frozen pre-trained model processes input sequences (truncated to a manageable Pattern Capture Length, PCL) to extract full attention maps. These maps undergo a Box-Cox transformation for normalization, followed by thresholding to generate "true masks" representing key dependencies. Structural pattern analysis (identifying vertical and diagonal patterns) then enables the extrapolation of these masks to lengths exceeding the PCL, creating "extended masks". The Appendix \ref{appx:attn_patterns} describes the detailed extension observation.

The second stage applies these generated, adaptive sparse attention masks (either true or extended, depending on sequence length) to a sparse Transformer model. This application occurs \textit{before} the softmax operation within the attention mechanism, effectively limiting computations to the unmasked connections. This significantly reduces both memory and computational overhead compared to full attention, while preserving the crucial dependencies identified in the first stage. By focusing on observed attention patterns and extrapolating structural regularities, \modelname achieves a balance between efficiency and the ability to capture nuanced, long-range dependencies in input sequences, making it suitable for processing long sequences that would otherwise be computationally prohibitive.

%%%%%%%%%%%%%%%%%%%%%%%

\subsubsection{Pattern Capture Length (PCL)}\label{sec:length_define}

The Pattern Capture Length (PCL), denoted as \(L\), represents a critical parameter within the \modelname framework. It defines the maximum sequence length processed by the \textit{frozen} model to extract the initial, full attention distributions. This constraint is essential for maintaining computational feasibility, particularly given the quadratic complexity of full attention mechanisms.

Let \(S\) represent the length of an input sequence. The PCL, \(L\), is determined as $L = \min(S, L_{\max})$ where \(L_{\max}\) is the maximum sequence length for which full attention computation remains computationally tractable given the available resources (e.g., GPU memory). 
As shown in Table \ref{tab:efficiency_exp}, the original LLaMA 3.2 3B model runs out of memory (OOM) when processing sequences longer than 8k tokens on an A100 GPU (40GB). Based on this, we \textbf{select the longest sequence length that the hardware can stably support}, and then adjust downward only if necessary. Unlike tuning from small to large values, which is inefficient and error-prone, starting from the maximum supported length and adjusting downward as needed makes PCL tuning both simple and reliable.

In essence, the PCL acts as a truncation point, ensuring that the initial attention map extraction is performed on sequences of a manageable length, while still capturing representative attention patterns. The choice of \(L_{\max}\) is a hyperparameter that depends on the specific hardware and model architecture.

\subsubsection{Feature Amplification via Box-Cox}\label{sec:feature_amplification}

This section details the process of amplifying and normalizing attention scores using the Box-Cox transformation. This step aims to address the often-observed skewness in attention distributions, where a few connections dominate while many others have very low values. By amplifying smaller attention values, we reveal potentially significant connections that might otherwise remain masked.

First, mean attention scores are computed across all valid position pairs within the dataset.  Let \( A_{\ell, h, i, j} \) denote the accumulated attention value at layer \( \ell \), head \( h \), from token position \( i \) to token position \( j \), summed across multiple batches. A binary mask \( m_{i, j}^{(b)} \in \{0, 1\} \) indicates whether the attention weight for position pair \((i, j)\) was computed in batch \( b \).  The count matrix \( C_{\ell, h, i, j} \) records the number of times each position pair \((i,j)\) appears across all batches, we have $C_{\ell, h, i, j} = \sum_{b} m_{i, j}^{(b)}$. The mean attention score, \(\bar{A}_{\ell, h, i, j}\), is then calculated as:

\begin{equation*}
\bar{A}_{\ell, h, i, j} = \frac{A_{\ell, h, i, j}}{C_{\ell, h, i, j} + \epsilon},
\end{equation*}

\noindent
where \(\epsilon\) is a small constant (e.g., \(10^{-8}\)) added to the denominator to prevent division by zero and ensure numerical stability.

To mitigate the skewness of the attention scores and emphasize smaller values, a Box-Cox transformation is applied. To ensure the input to the transformation is strictly positive, a small constant \(\epsilon\) is added to the mean attention scores as $X_{\ell, h, i, j} = \max(\bar{A}_{\ell, h, i, j}, \epsilon)$. The Box-Cox transformation is then applied to \(X_{\ell, h, i, j}\) as follows:

\begin{equation*}
B_{\ell, h, i, j} =
\begin{cases}
\frac{X_{\ell, h, i, j}^\lambda - 1}{\lambda}, & \text{if } \lambda \neq 0 \\
\ln(X_{\ell, h, i, j}), & \text{if } \lambda = 0
\end{cases}
\end{equation*}

\noindent
where \(\lambda\) is the transformation parameter. In practice, we find that \(\lambda = 0.5\) improves visualization, which does not require to be tuned in the future.

To ensure the transformed values \(B_{\ell, h, i, j}\) remain non-negative, we subtract the minimum value across all heads and layers:

\begin{equation*}
B_{\ell, h, i, j}^* = B_{\ell, h, i, j} - \min_{\ell', h', i', j'}(B_{\ell', h', i', j'}).
\vspace{-0.4cm}
\end{equation*}

Finally, the normalized attention map \(\tilde{A}_{\ell, h, i, j}\) is defined as $\tilde{A}_{\ell, h, i, j} = B_{\ell, h, i, j}^*$. With amplified smaller values, \(\tilde{A}_{\ell, h, i, j}\) is then used for subsequent mask generation.

\subsubsection{True Mask Generation}\label{sec:true_mask}

We detail the process of generating "true masks", denoted as  \(M_{\ell, h}\), which represent the binarized and thresholded version of the normalized attention maps. These masks serve as the basis for identifying structural patterns and subsequently constructing the extended, sparse attention masks.

A binary thresholding operation is applied to the normalized attention maps, \(\tilde{A}_{\ell, h}\) (obtained as described in Section~\ref{sec:feature_amplification}), to produce the true masks.  Each true mask \(M_{\ell, h}\) has the same dimensions as the corresponding attention map: $M_{\ell, h} = \left[ m_{i,j} \right] \in \{0,1\}^{L \times L}$, where \(L\) is the Pattern Capture Length (PCL). The elements of the true mask, \(m_{i,j}\), are determined by comparing the corresponding normalized attention values, \(\tilde{A}_{\ell, h, i, j}\), to a predefined threshold, \(\tau\):
% \vspace{-0.2cm}

\begin{equation*}
m_{i,j} =
\begin{cases}
1, & \text{if } \tilde{A}_{\ell, h, i, j} \geq \tau, \\
0, & \text{if } \tilde{A}_{\ell, h, i, j} < \tau.
\end{cases}
\end{equation*}

This thresholding operation is applied independently to each layer \(\ell\) and attention head \(h\). The threshold, \(\tau\), acts as a hyperparameter controlling the sparsity of the true masks. A higher value of \(\tau\) results in a sparser mask, retaining only the strongest attention connections.

\subsubsection{Dynamic Mask Generation via Structural Pattern Matching}\label{sec:mask_generation}

We construct \modelname by identifying and combining predefined structural patterns within the true attention masks. 
A pattern pool, \(\mathcal{P}\), is defined, consisting of a set of predefined attention patterns. Each pattern is represented as a binary matrix $P_k = \left[ p_{i,j} \right] \in \{0,1\}^{L \times L}$, where \(L\) denotes the PCL and  \(P_k\) represents the \(k\)-th pattern in the pool. The pattern pool, in this work, includes diagonal and vertical patterns, reflecting common attention structures observed in Transformer models.

A diagonal pattern, \(P_{\text{diag}, r}\), starts at row index \(r\) and extends diagonally downwards:

\begin{equation*}
p_{i,j} =
\begin{cases}
1, & \text{if } j = i - r, \\ 
0, & \text{otherwise}.
\end{cases}
\end{equation*}

\noindent
for \( r \in \{0, 1, \dots, L-1\} \). A vertical pattern, \(P_{\text{vert}, c}\), captures column-wise attention (i.e., tokens attending to a specific column \(c\)):

\begin{equation*}
p_{i,j} =
\begin{cases}
1, & \text{if } j = c \text{ and } i \geq c, \\ 
0, & \text{otherwise}.
\end{cases}
\end{equation*}

\noindent
for \( c \in \{0, 1, \dots, L-1\} \). The complete pattern pool is the union of these sets:

\begin{equation*}
\mathcal{P} = \{ P_{\text{diag}, r} \} \cup \{ P_{\text{vert}, c} \}.
\end{equation*}

Each true mask \( M_{\ell, h} \) is compared against patterns in \( \mathcal{P} \). The match score \( \gamma_k \) for a pattern \( P_k \) is computed as:
\vspace{-0.3cm}

\begin{equation*}
\gamma_k = \frac{\sum_{i,j} M_{\ell, h}^{(i,j)} \cdot P_k^{(i,j)}}{\sum_{i,j} P_k^{(i,j)}}.
\end{equation*}

A pattern \(P_k\) is considered a valid match if its match score \(\gamma_k\) exceeds a predefined threshold \(\mu\), where \(\mu \in [0, 1]\) is a hyperparameter controlling the sensitivity of the pattern matching. Higher values of \(\mu\) lead to fewer patterns being matched, resulting in sparser masks. It is robust across a relatively wide range, from 0.7 to 1.0, while still preserving the model's language understanding capabilities.

The extended mask, \(\tilde{M}_{\ell, h}\), is constructed by extrapolating the structural patterns identified in the true masks, which are initially computed using sequences up to the PCL. These patterns—such as diagonal and vertical structures—are selected from a predefined pattern pool. Because patterns may overlap, the extended mask is formed by summing all matched patterns whose match score exceeds a threshold:

\begin{equation*}
\tilde{M}_{\ell, h} = \sum_{P_k \in \mathcal{P}, \gamma_k \geq \mu} P_k.
\end{equation*}

Finally, to ensure the extended mask is binary, a thresholding operation is applied:

\begin{equation*}
\tilde{M}_{\ell, h}^{(i,j)} =
\begin{cases}
1, & \text{if } \sum_{P_k \in \mathcal{P}, \gamma_k \geq \mu} P_k^{(i,j)} \geq 1, \\ 
0, & \text{otherwise}.
\end{cases}
\end{equation*}

\subsection{Applying Dynamic Attention Masks}\label{sec:applying_masks}

\textbf{Case 1:} If the input sequence length \( S \) satisfies \( S \leq L \), \( M_{\ell, h} \) will be applied as the attention mask.

\noindent
\textbf{Case 2:} If \( S > L \), the method constructs an extended mask \( \tilde{M}_{\ell, h} \) of size \( S \times S \). The first \( L \times L \) region remains unchanged:

\begin{equation*}
\tilde{M}_{\ell, h}^{(i,j)} = M_{\ell, h}^{(i,j)}, \quad \text{for } i,j \leq L.
\end{equation*}

For \( i, j > L \), attention is allowed if the token pair \( (i, j) \) is in the stored matched positions \( \mathcal{P}_{\ell, h} \):

\begin{equation*}
\tilde{M}_{\ell, h}^{(i,j)} =
\begin{cases} 
1, & \text{if } (i,j) \in \mathcal{P}_{\ell, h}, \\ 
0, & \text{otherwise}.
\end{cases}
\end{equation*}

The attention mask applies before softmax. The modified attention score matrix is:

\begin{equation*}
A'_{\ell, h} = \frac{Q_{\ell, h} K_{\ell, h}^T}{\sqrt{d_k}} \odot \tilde{M}_{\ell, h}.
\end{equation*}

The model sets masked positions \( \tilde{M}_{\ell, h}^{(i,j)} = 0 \) to \( -\infty \) before softmax, ensuring a probability of zero. The final output is: $O'_{\ell, h} = \text{softmax}(A'_{\ell, h}) V_{\ell, h}$.

%%%%%%%%%%%%%%%%%%%%%%%%%%%%%%%%%%%%%%%%%%%%%%%%%%%%%%%%%%%%%%
%%%%%%%%%%%%%%%%%%%%%%%%%%%%%%%%%%%%%%%%%%%%%%%%%%%%%%%%%%%%%%

\section{Experiment}

\subsection{Experiment Setup}

\begin{figure}[htbp]
    \centering
    \vspace{-0.1cm}
    \includegraphics[width=1\linewidth]{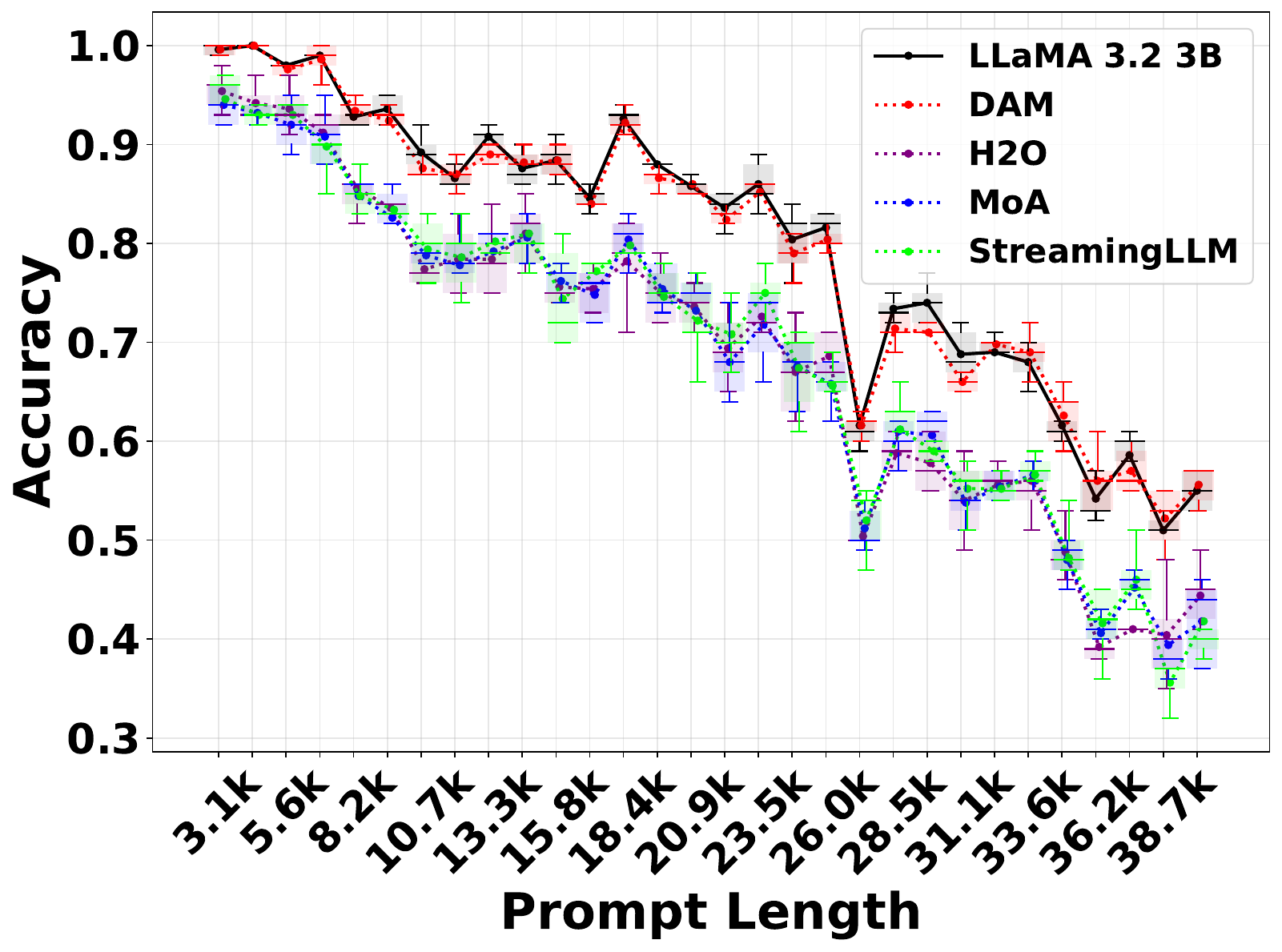}
    \vspace{-0.6cm}
    \caption{Line-level retrieval accuracy on the LongEval benchmark using the base LLaMA 3.2 3B model. Each sequence contains a predefined target line to be retrieved. Input lengths range from 200 to 3100 lines (3.1k to 38.7k tokens). DAM maintains high retrieval accuracy across all lengths with minimal degradation. }
    \label{fig:LongEval_length}
    \vspace{-0.3cm}
\end{figure}

We evaluate DAM on long-context retrieval and QA tasks, comparing against full attention and structured sparsity baselines across multiple sequence lengths and model scales.

\noindent
\textbf{Baselines.} We compare \modelname against FlashAttention~\cite{dao2023flashattention}, MoA~\cite{fu2024moa}, StreamingLLM~\cite{xiao2024efficientstreaminglanguagemodels}, and H2O~\cite{NEURIPS2023_6ceefa7b}. MoA uses predefined sparse attention patterns per layer and head, while StreamingLLM and H2O enhance efficiency during autoregressive decoding.

\noindent
\textbf{Base Models.} The experiments use LLaMA-3.2-1B-Instruct and LLaMA-3.2-3B-Instruct to analyze scalability across different parameter sizes. 

\noindent
\textbf{Benchmarks.} The evaluation uses LongEval~\cite{krishna2023longeval} and LV-Eval~\cite{yuan2024lvevalbalancedlongcontextbenchmark} to assess long-context understanding. LongEval measures key-value retrieval accuracy with 100 data items per sequence length level, offering insights into contextual recall performance.

\noindent
\textbf{Hardware.} The experiments run on multiple GPU configurations: 4 $\times$ A100 (40GB) for LongEval, 2 $\times$ H100 (80GB) for LV-Eval, and 1 $\times$ A100 (40GB) for efficiency evaluations.

% \noindent
\textbf{DAM Configuration:}  
\begin{itemize}[nosep,left=0pt]
    \item \textbf{Dataset for attention map capture:} Multi-News~\cite{DBLP:journals/corr/abs-1906-01749}, a large-scale multi-document dataset that captures diverse attention patterns for general language capability.  
    \item \textbf{Pattern Capture Length:} 512, balancing feasibility with attention pattern extraction.  
    \item \textbf{Threshold for true masks:} 0.3, determined through attention sparsity analysis.  
    \item \textbf{Threshold for approximate masks:} 0.8, ensuring effective structural alignment while minimizing unnecessary attention connections.  
\end{itemize}

\begin{figure*}[ht]
    \centering
    \vspace{-0.3cm}
    \includegraphics[width=1\linewidth]{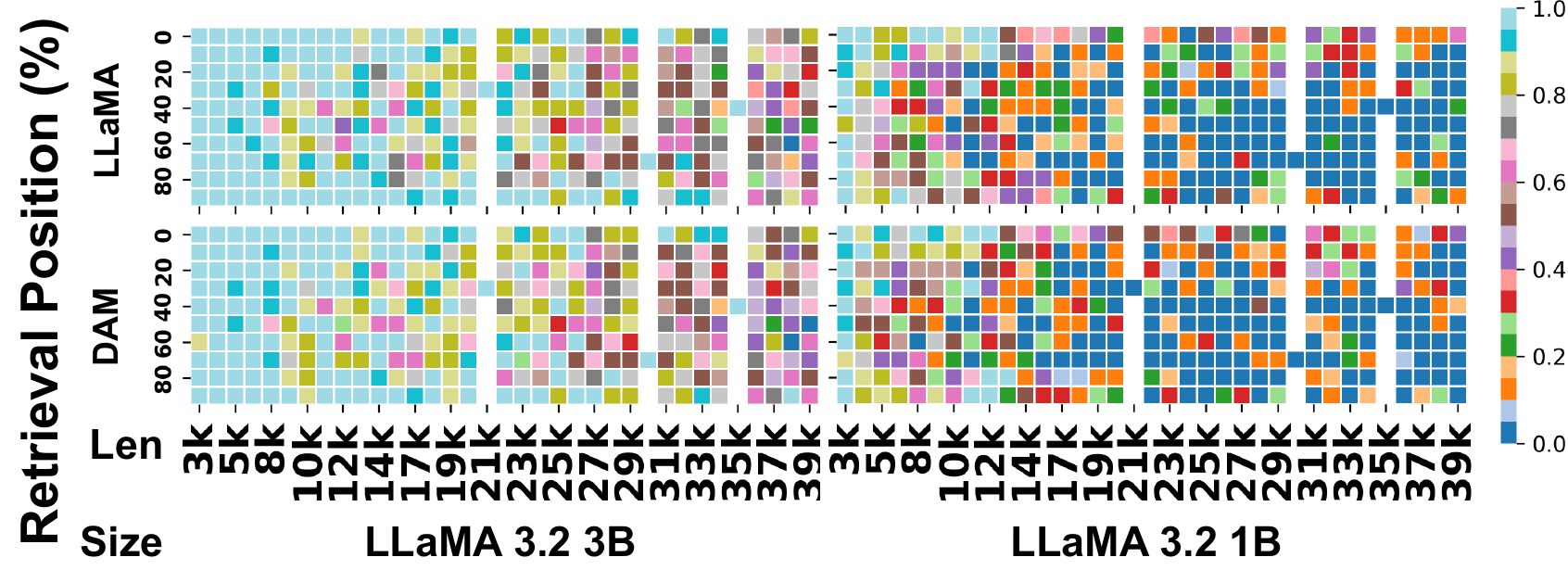}
    \vspace{-0.6cm}
    \caption{Retrieval accuracy on LongEval for LLaMA 3.2 3B and 1B models. Even at fixed token lengths, performance varies based on the target keyword’s position within the sequence. DAM closely matches the dense model’s retrieval accuracy across all settings.}
    \label{fig:LongEval_retrieve}
\end{figure*}

\begin{figure*}[t]
    \centering
    \includegraphics[width=\textwidth]{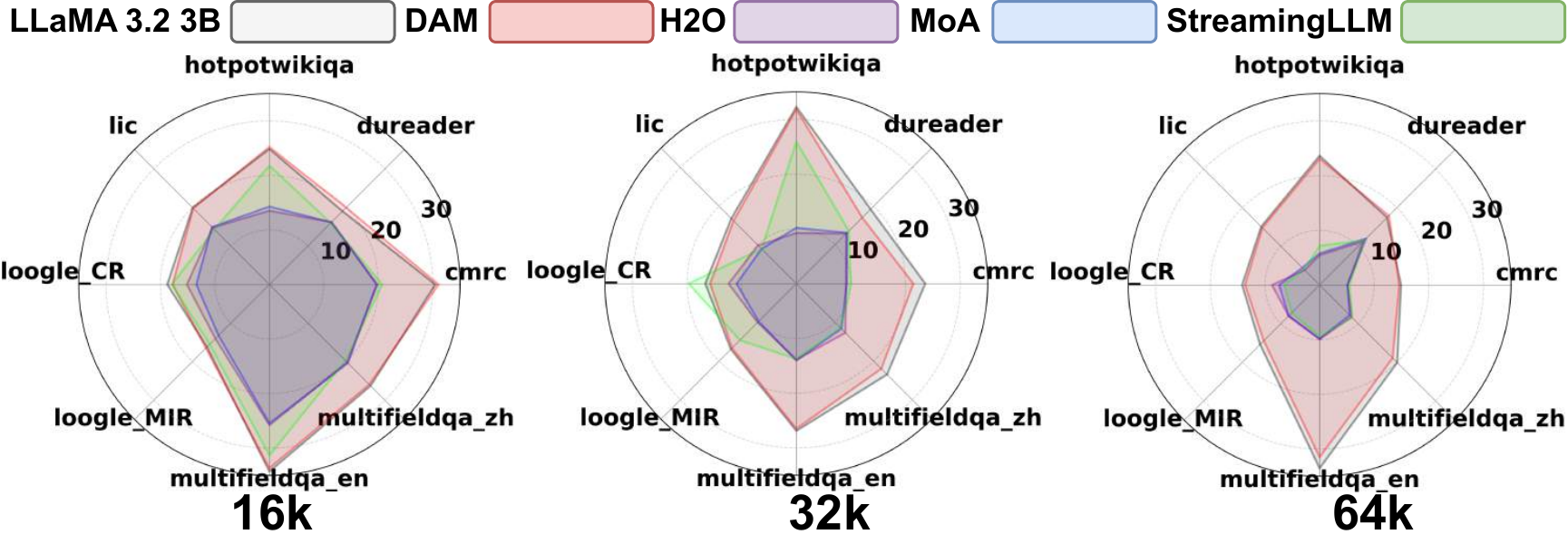}
    \vspace{-0.6cm}
    \caption{LV-Eval retrieval score across long-context QA tasks. DAM closely matches full attention, achieving 18.61 score at 64K tokens. MoA, StreamingLLM, and H2O lose performance as sequence length increases, with DAM outperforming alternative sparse attention methods.}
    \label{fig:LVEval}
    \vspace{-0.4cm}
\end{figure*}

\subsection{Performance Evaluation}

\textbf{Long-Context Retrieval.} The LongEval lines task evaluates retrieval accuracy across different sequence lengths by measuring a model’s ability to extract predefined tokens embedded within input sequences ranging from 3K to 104K tokens (illustration ends with base model accuracy smaller than 0.5). 
Figure~\ref{fig:LongEval_length} shows that DAM maintains an average accuracy of 0.7966, closely matching full attention (0.8011). The accuracy gap remains minimal across all tested lengths, confirming DAM’s ability to preserve long-range dependencies.
MoA and StreamingLLM experience sharp performance declines beyond 20K tokens, with accuracy dropping to 0.394 and 0.356, respectively. 
These models fail to capture heterogeneous attention patterns dynamically, leading to reduced retrieval accuracy. 

The retrieval accuracy task evaluates the ability to locate predefined tokens across different sequence lengths. Figure~\ref{fig:LongEval_retrieve} illustrates the performance of DAM compared to the full attention baseline on LLaMA 3.2 1B and 3B models. DAM consistently aligns with the dense model’s performance across all evaluated sequence lengths and keyword positions. Notably, even at the same token length, retrieval accuracy varies depending on the keyword’s relative position, highlighting the importance of modeling position-sensitive dependencies. DAM preserves such fine-grained retrieval capabilities, demonstrating its effectiveness at retaining long-range and position-sensitive attention patterns without incurring the full computational cost of dense attention. The Appendix \ref{appx:LongEval_all} shows the full comparison of other methods.

\textbf{Long-Context Tasks.} The LV-Eval benchmark evaluates retrieval performance in long-context question-answering tasks. This experiment examines sequence lengths from 16K to 256K tokens, with results showing up to 64K tokens. Beyond 128K tokens, base model performance remains stable, while retrieval score declines at 256K tokens. The benchmark includes single-hop and multi-hop QA tasks that require retrieving relevant information from long input contexts.
Single-hop QA datasets include cmrc-mixup, multifieldqa-en-mixup, and multifieldqa-zh-mixup. Multi-hop QA datasets include dureader-mixup, loogle-CR-mixup, loogle-MR-mixup, hotpotwikiqa-mixup, and lic-mixup.

Figure~\ref{fig:LVEval} shows that DAM closely follows full attention across all datasets. At 64K tokens, DAM reaches an average score of 18.61, compared to 19.29 for full attention. The small gap confirms DAM’s ability to retain retrieval scores without quadratic attention costs.
MoA and StreamingLLM lose scores as sequence length increases. At 64K tokens, MoA reaches 7.56 and StreamingLLM 7.47, both lower than DAM and full attention. These models fail to retain long-range dependencies, reducing effectiveness in multi-hop retrieval.
H2O holds performance better than MoA and StreamingLLM but scores lower than DAM. At 64K tokens, H2O records 7.59, slightly above MoA and StreamingLLM but below DAM. 

\begin{table}[t]
  \centering
  \renewcommand{\arraystretch}{0.9}   
  \setlength{\tabcolsep}{3pt}         
  \begin{tabularx}{\linewidth}{c|c|c|c|c|c}
    \toprule
    \textbf{Mdl} 
      & \textbf{Method} 
        & \textbf{Len} 
          & \textbf{Mem} 
            & \textbf{Tupt} 
              & \textbf{AvgTim} \\
    \midrule
    \multirow{12}{*}{\rotatebox{90}{LLaMA 3.2 1B}}
      & \multirow{4}{*}{Original}
          & 1k &  5.21   & 1677.63  &  9766.18   \\
      &                         & 2k & 12.13   & 1025.84  & 31942.5    \\
      &                         & 4k & 38.10   &  928.18  & 70607      \\
      &                         & 8k & OOM     & —        & —          \\
    \cmidrule(lr){2-6}
      & \multirow{4}{*}{FlashAttn}
          & 1k &  3.82   & 1763.55  &  9290.38   \\
      &                         & 2k &  5.32   & 1099.35  & 29806.8    \\
      &                         & 4k &  8.33   &  633.456 & 103458     \\
      &                         & 8k & 10.21   & 69823.4  &  1877.19   \\
    \cmidrule(lr){2-6}
      & \multirow{4}{*}{\textbf{DAM}}
          & 1k &  3.84   & 2574.84  &  6363.11   \\
      &                         & 2k &  5.35   & 1656.09  & 19786.4    \\
      &                         & 4k &  8.35   &  941.22  & 69628.8    \\
      &                         & 8k & 10.64   &  639.653 & 204911     \\
    \midrule
    \multirow{12}{*}{\rotatebox{90}{LLaMA 3.2 3B}}
      & \multirow{4}{*}{Original}
          & 1k &  9.89   &  689.20  & 23772.5    \\
      &                         & 2k & 16.70   &  403.58  & 81194.1    \\
      &                         & 4k & OOM     & —        & —          \\
      &                         & 8k & OOM     & —        & —          \\
    \cmidrule(lr){2-6}
      & \multirow{4}{*}{FlashAttn}
          & 1k &  9.88   &  710.65  & 23055.1    \\
      &                         & 2k & 13.76   &  419.17  & 78173.8    \\
      &                         & 4k & 21.52   &  232.15  & 282298     \\
      &                         & 8k & 21.15   & 25796.4  &  5081.03   \\
    \cmidrule(lr){2-6}
      & \multirow{4}{*}{\textbf{DAM}}
          & 1k &  9.90   & 1095.52  & 14955.4    \\
      &                         & 2k & 13.79   &  651.14  & 50324.3    \\
      &                         & 4k & 21.53   &  354.96  & 184631     \\
      &                         & 8k & 31.71   &  238.08  & 550541     \\
    \midrule
    \multirow{4}{*}{\rotatebox{90}{Vicuna 7B}}
      & \multirow{2}{*}{Original}
          & 1k & 28.81   &  451.42  & 36294.7    \\
      &                         & 2k & OOM     & —        & —          \\
    \cmidrule(lr){2-6}
      & \multirow{2}{*}{\textbf{DAM}}
          & 1k & 28.84   &  738.97  & 22171.5    \\
      &                         & 2k & 39.14   &  437.05  & 74976      \\
    \bottomrule
  \end{tabularx}
  \caption{Comparison of GPU memory (GB), throughput (tokens/s), and average latency (ms) for Original, FlashAttention, and DAM across LLaMA 3.2 (1B, 3B) and Vicuna 7B models at various sequence lengths.}
  \label{tab:efficiency_exp}
  \vspace{-0.6cm}
\end{table}

\subsection{Efficiency}

We benchmark DAM against the original dense attention implementation and FlashAttention across LLaMA 3.2 1B, 3B, and Vicuna 7B models. Table~\ref{tab:efficiency_exp} reports GPU memory usage (GB), average decoding latency (ms), and throughput (tokens/sec) at varying sequence lengths.

DAM consistently maintains a low memory footprint across all models and sequence lengths. For LLaMA 3.2 1B, DAM uses only 10.64 GB at 8K tokens, compared to 38.1 GB for the dense model (which fails beyond 8K due to OOM). Similarly, for LLaMA 3.2 3B and Vicuna 7B, DAM enables 8K inference while full attention is infeasible. DAM’s structured sparsity yields memory savings comparable to FlashAttention, and enables longer context handling without modifying model weights or kernel implementations.

DAM achieves favorable throughput–latency trade-offs compared to FlashAttention. For LLaMA 1B at 4K tokens, DAM reaches 941 tokens/sec (vs. 633 for FlashAttention) with 33\% lower latency. At 8K, DAM maintains 639 tokens/sec with 204 ms latency, whereas FlashAttention achieves a higher 69K tokens/sec spike but only due to GPU tiling effects. This discrepancy stems from FlashAttention’s recompute-based kernel optimization, which exhibits nonlinear scaling when the input length aligns with block sizes (e.g., 8192 = 64×128). In contrast, DAM’s performance scales more predictably across lengths, without relying on such alignment.

FlashAttention and DAM operate at different layers of optimization. FlashAttention optimizes kernel-level memory access for dense attention, while DAM applies structured sparsity at the model level by preselecting important attention positions. Although we compute full attention maps initially (for mask extraction), DAM avoids attending to most token pairs during inference. This reduces FLOPs complexity from $\mathcal{O}(L^2)$ to $\mathcal{O}(sL)$, where $s$ is the average number of retained keys per query ($s \ll L$). Importantly, DAM’s sparse attention layout is compatible with tile-based GPU execution, allowing further fusion with memory-efficient kernels like FlashAttention in future implementations.

%%%%%%%%%%%%%%%%%%%%%%%%%%%%%%%%%%%%%%%%%%%%%%%%%%%%%%%%%%%%%%
%%%%%%%%%%%%%%%%%%%%%%%%%%%%%%%%%%%%%%%%%%%%%%%%%%%%%%%%%%%%%%
\section{Conclusion}
We introduce \modelname, a sparse attention method that dynamically captures heterogeneous token interactions, overcoming the limitations of static and predefined sparsity patterns. \modelname learns adaptive attention masks that retain crucial dependencies, improving retrieval accuracy while significantly reducing computational cost. 
Experiments across long-context benchmarks demonstrate DAM’s effectiveness in maintaining full-attention performance with lower memory and compute requirements. By bridging the gap between efficiency and expressivity in sparse attention, DAM provides a scalable solution for long-context processing.

\newpage
\section{Limitations}
While reducing attention computation at runtime, dynamic sparse masks add preprocessing overhead versus fixed masks. 
Optimizing mask generation to minimize overhead while maintaining adaptability remains a challenge.
Additionally, the approach assumes that structured sparsity in attention patterns can be effectively learned and generalized, but this may not always align with optimal information flow in every task. 
Future work could explore adaptive learning mechanisms that refine sparsity patterns based on downstream task performance.
Though this method scales more efficiently than full attention, handling extremely long sequences, such as multi-million-token documents or continuous streaming inputs, remains a challenge due to memory constraints in mask storage and extension. 
Exploring hybrid models that integrate retrieval-based or memory-augmented techniques could improve efficiency for such cases.

\bibliography{reference}

\newpage

\appendix

\section{Attention Pattern Observation}\label{appx:attn_patterns}

\begin{figure}[h]
    \centering
    \includegraphics[width=\linewidth]{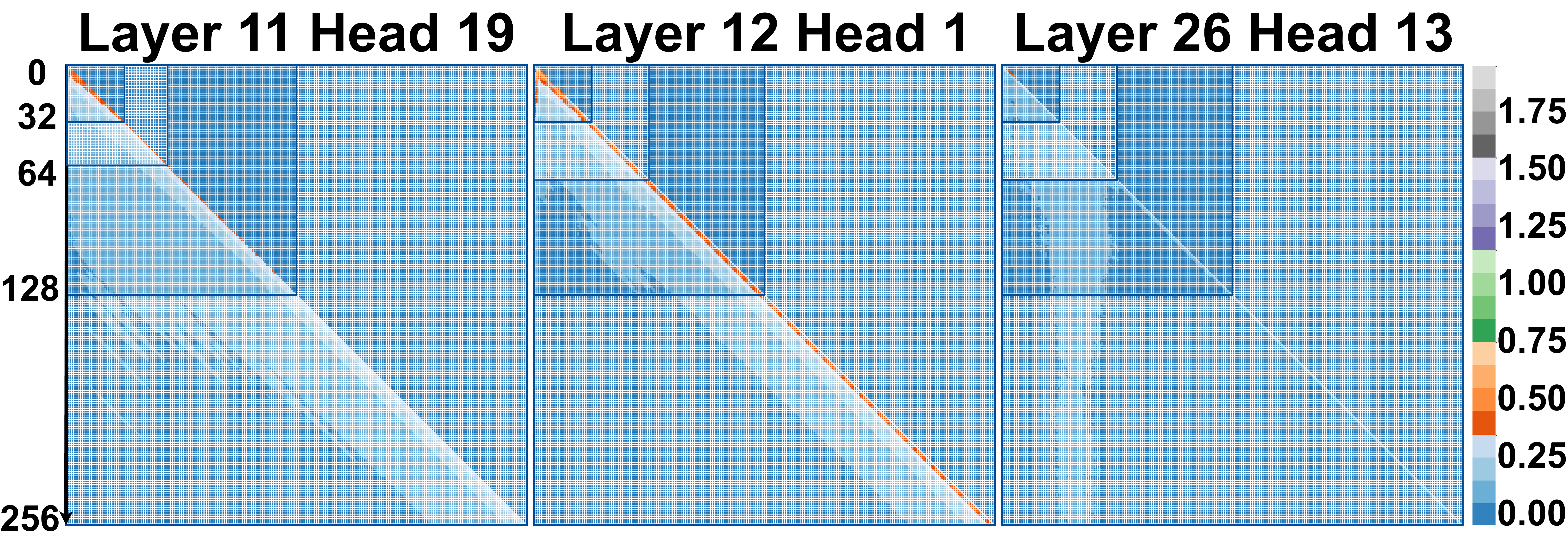}
    \caption{Attention patterns from the LLaMA 3.2 3B model across 32, 64, 128, 256 token lengths. In the left maps, sliding window patterns persist at 64 tokens but disappear by 256. The middle maps show transient tilt patterns from 64 to 128 that do not form regular shapes. Longer token length attention maps extend shorter ones at the same layer and head.}
    \label{fig:attention_patterns}
\end{figure}

Figure~\ref{fig:attention_patterns} shows how attention patterns evolve with increasing sequence length in Layer 11 Head 19, Layer 12 Head 1, and Layer 26 Head 13 of the LLaMA 3.2 3B model.

Layer 11 Head 19 (left) exhibits a clear sliding window structure at short lengths (32–64 tokens), which gradually fades by 256 tokens. Layer 12 Head 1 (middle) displays irregular diagonal tilts at intermediate lengths (64–128), though these patterns are unstable. In contrast, Layer 26 Head 13 (right) forms consistent, vertically aligned sparse structures as input length increases, suggesting stable token selection in deeper layers.

These patterns motivate the design of \modelname's extended attention masks. We extract reliable structural motifs—such as diagonal and vertical bands—from short sequences and use them to extrapolate attention behavior at longer lengths. This avoids recomputing full attention maps while preserving meaningful structure.

\modelname computes true attention up to a user-defined \maxlength, captures key patterns from the resulting \(L \times L\) mask, and replicates them to extend the mask for longer inputs. The final mask is applied before the softmax to enforce structured sparsity during inference, reducing computation and memory while maintaining attention fidelity.

\section{Feature Amplification Transformation Methods}\label{appx:trans_methods}

\begin{figure*}[htbp]
  \centering
  \setlength{\tabcolsep}{2pt}
  \renewcommand{\arraystretch}{1.0}
  \begin{tabular}{|c|*{6}{c|}}
    \hline
      & \textbf{L1H8} & \textbf{L3H11} & \textbf{L5H19} & \textbf{L15H3} & \textbf{L25H9} & \textbf{L27H13} \\
    \hline\hline

    \rotatebox{90}{\textbf{raw-sum}}
      & \includegraphics[width=0.15\linewidth]{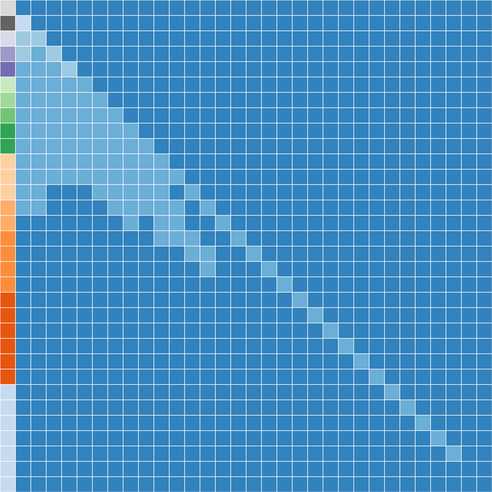}
      & \includegraphics[width=0.15\linewidth]{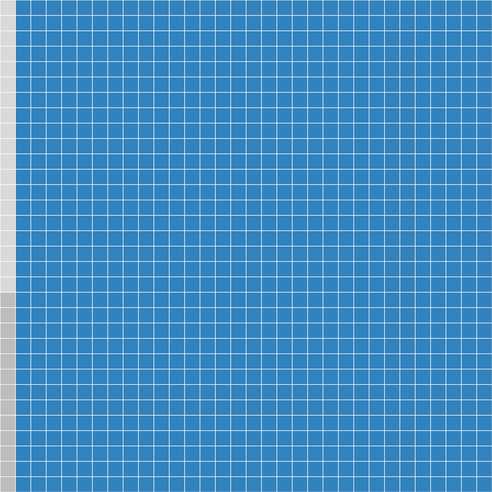}
      & \includegraphics[width=0.15\linewidth]{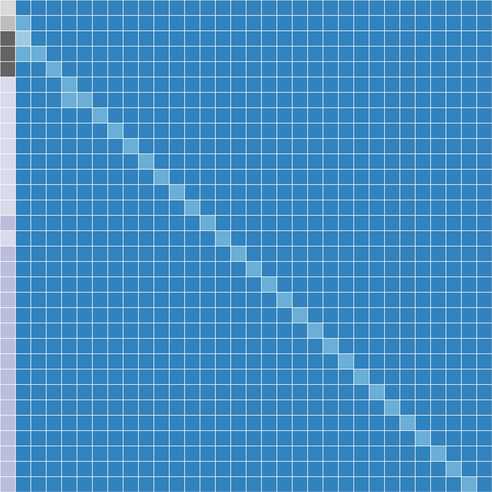}
      & \includegraphics[width=0.15\linewidth]{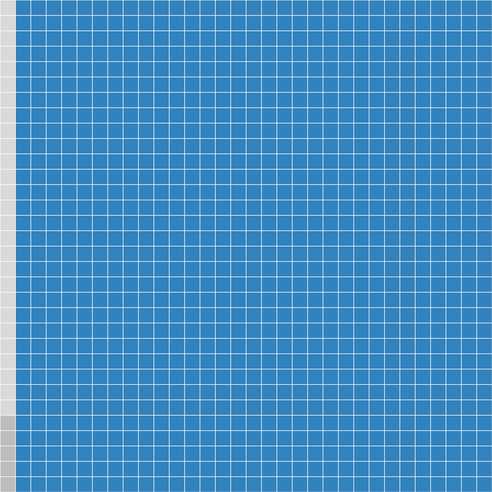}
      & \includegraphics[width=0.15\linewidth]{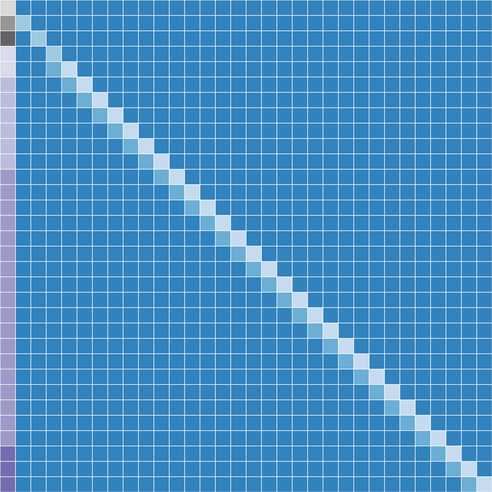}
      & \includegraphics[width=0.15\linewidth]{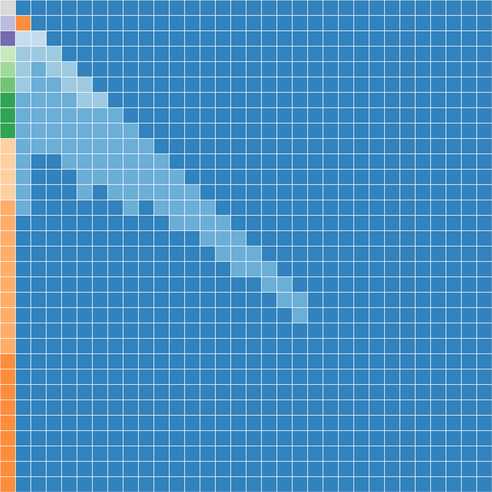} \\
    \hline

    \rotatebox{90}{\textbf{average}}
      & \includegraphics[width=0.15\linewidth]{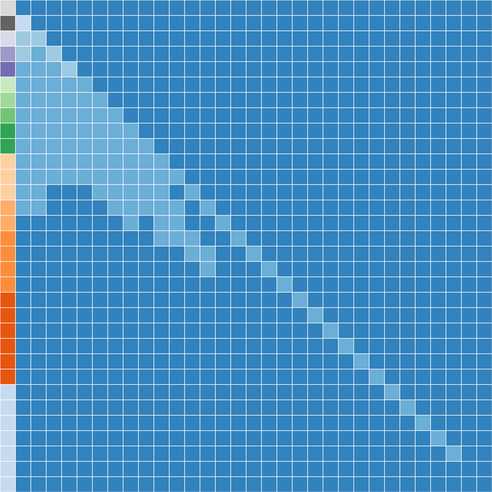}
      & \includegraphics[width=0.15\linewidth]{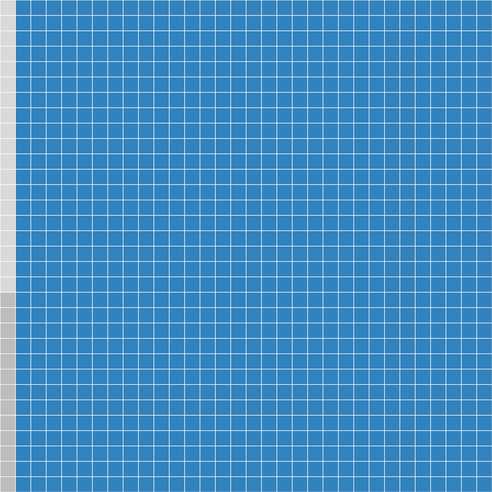}
      & \includegraphics[width=0.15\linewidth]{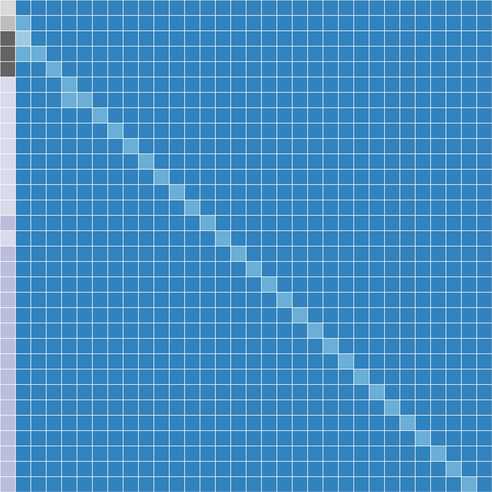}
      & \includegraphics[width=0.15\linewidth]{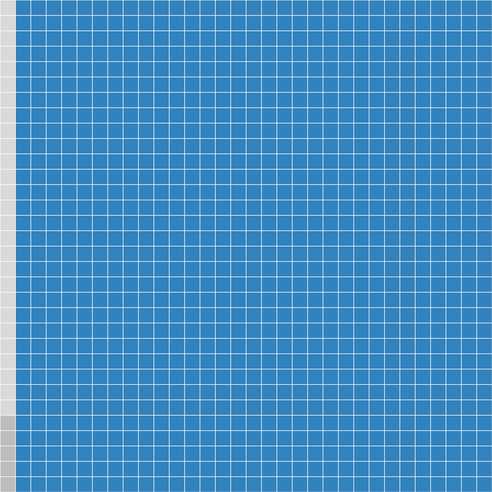}
      & \includegraphics[width=0.15\linewidth]{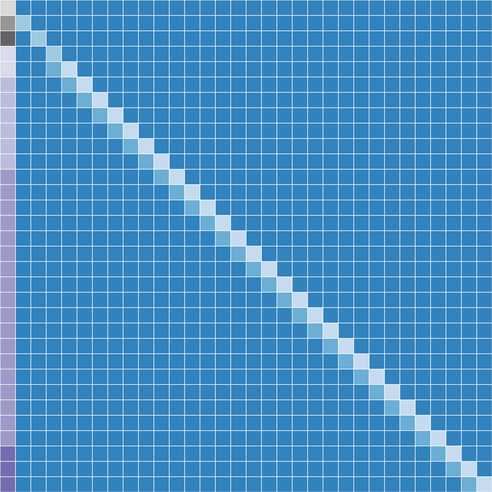}
      & \includegraphics[width=0.15\linewidth]{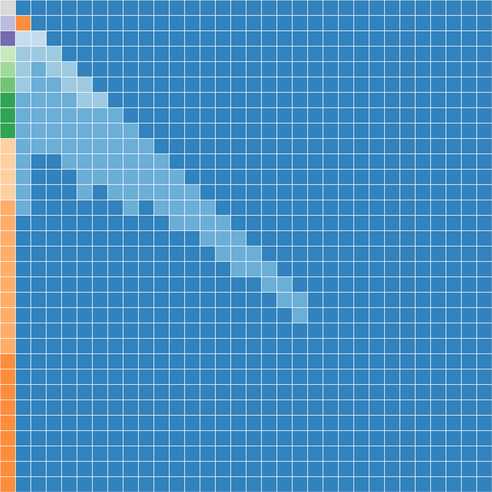} \\
    \hline

    \rotatebox{90}{\textbf{min-max}}
      & \includegraphics[width=0.15\linewidth]{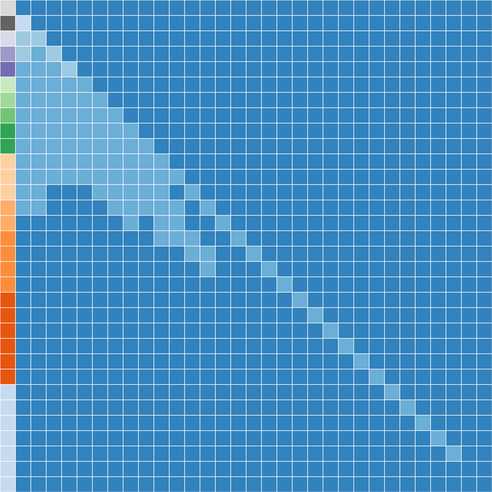}
      & \includegraphics[width=0.15\linewidth]{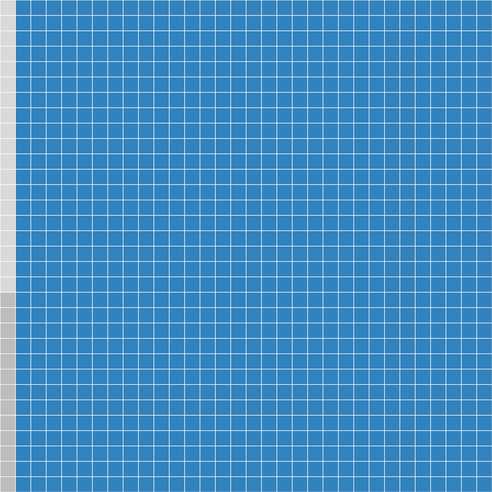}
      & \includegraphics[width=0.15\linewidth]{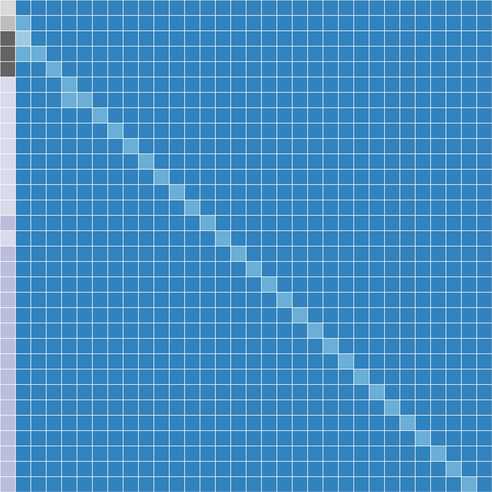}
      & \includegraphics[width=0.15\linewidth]{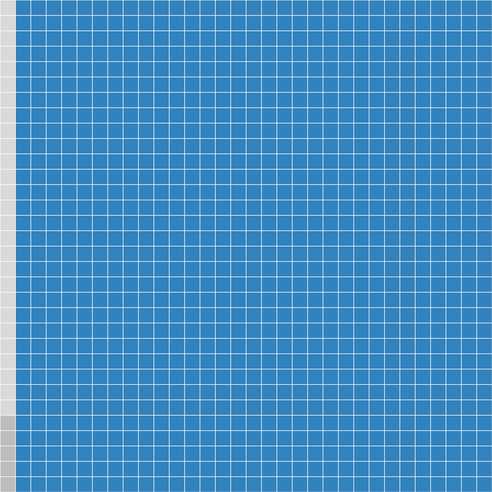}
      & \includegraphics[width=0.15\linewidth]{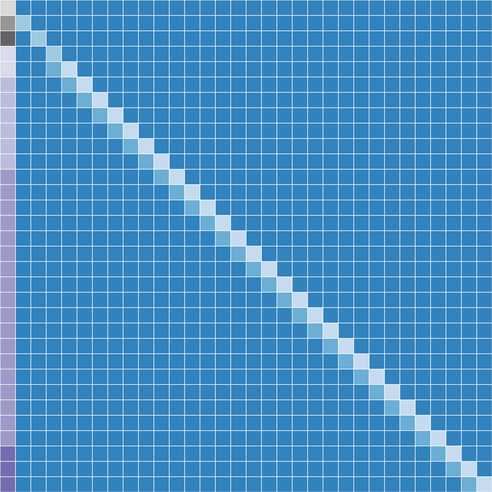}
      & \includegraphics[width=0.15\linewidth]{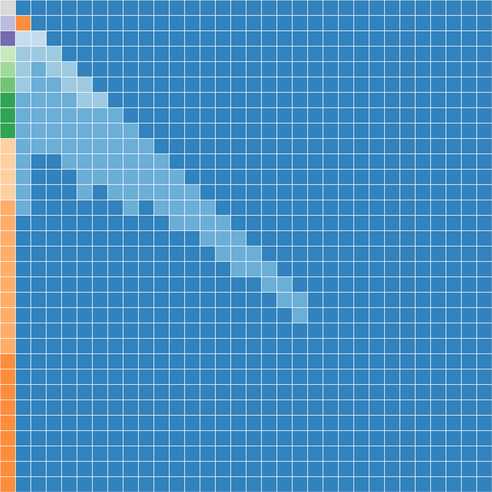} \\
    \hline

    \rotatebox{90}{\textbf{z-score}}
      & \includegraphics[width=0.15\linewidth]{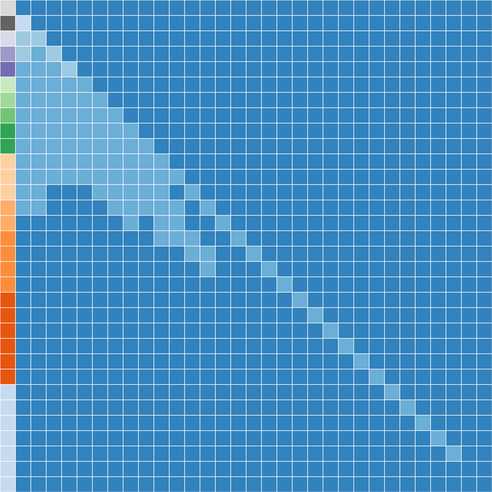}
      & \includegraphics[width=0.15\linewidth]{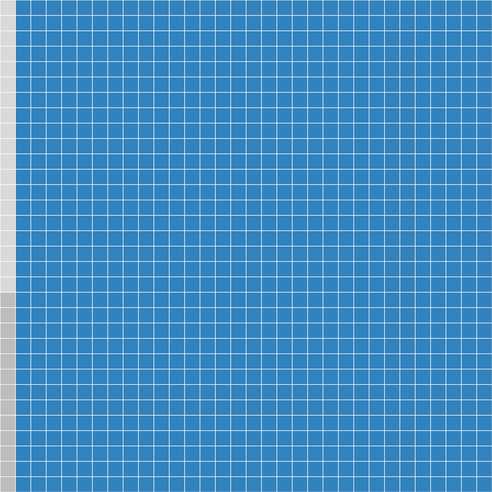}
      & \includegraphics[width=0.15\linewidth]{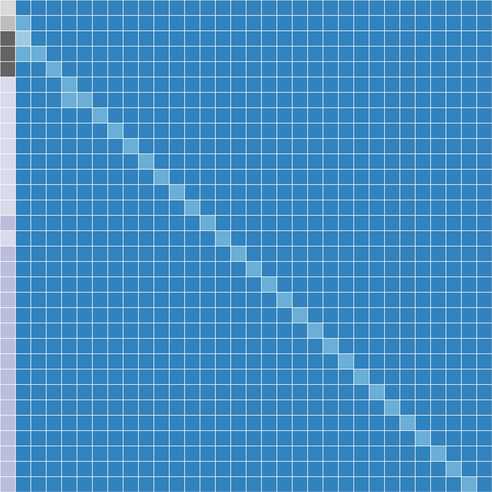}
      & \includegraphics[width=0.15\linewidth]{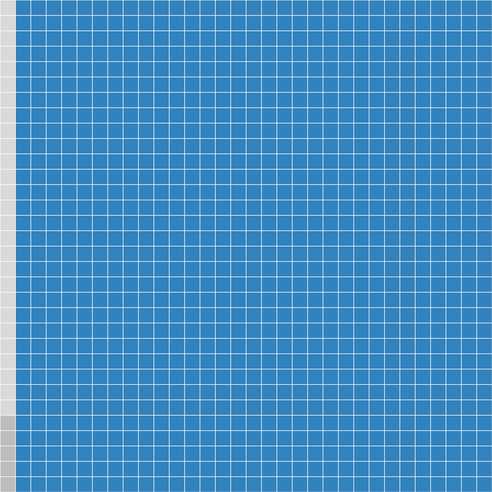}
      & \includegraphics[width=0.15\linewidth]{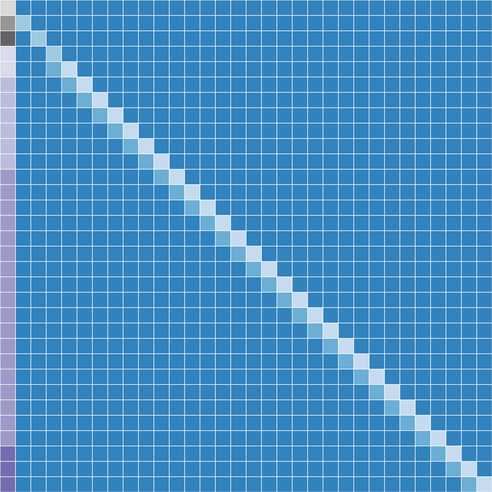}
      & \includegraphics[width=0.15\linewidth]{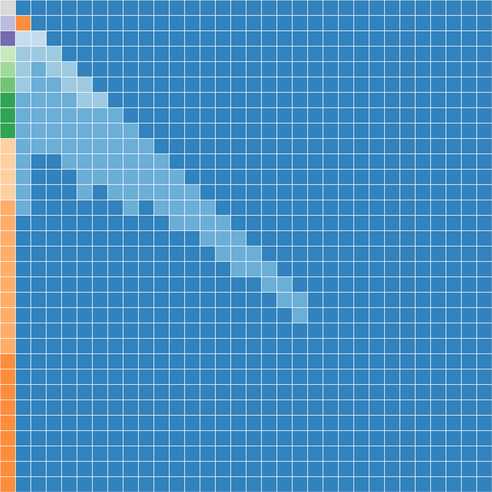} \\
    \hline

    \rotatebox{90}{\textbf{yeo-johnson}}
      & \includegraphics[width=0.15\linewidth]{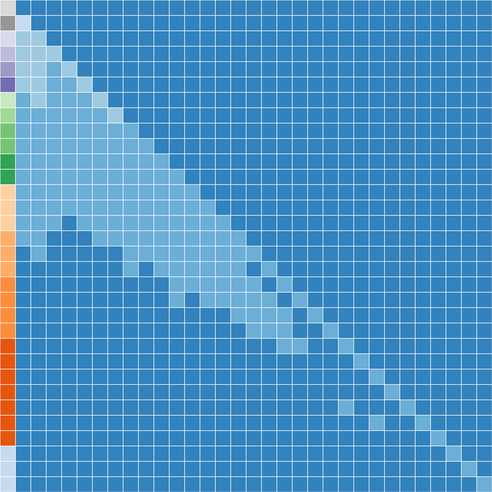}
      & \includegraphics[width=0.15\linewidth]{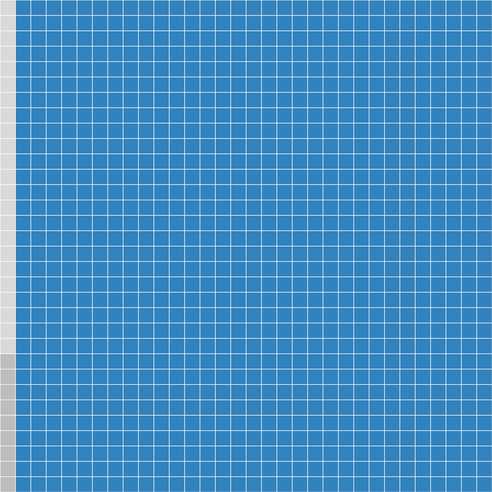}
      & \includegraphics[width=0.15\linewidth]{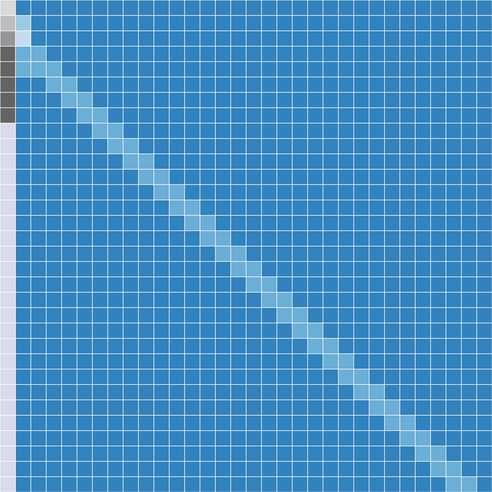}
      & \includegraphics[width=0.15\linewidth]{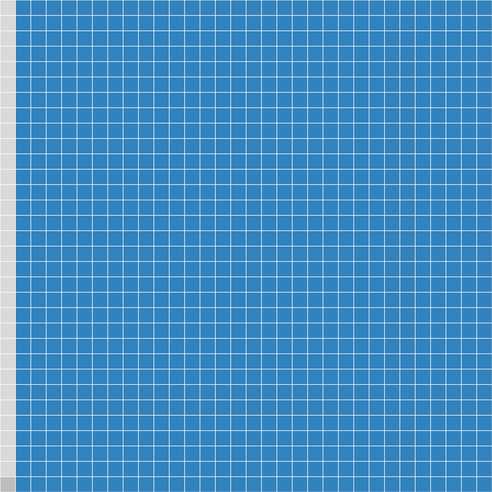}
      & \includegraphics[width=0.15\linewidth]{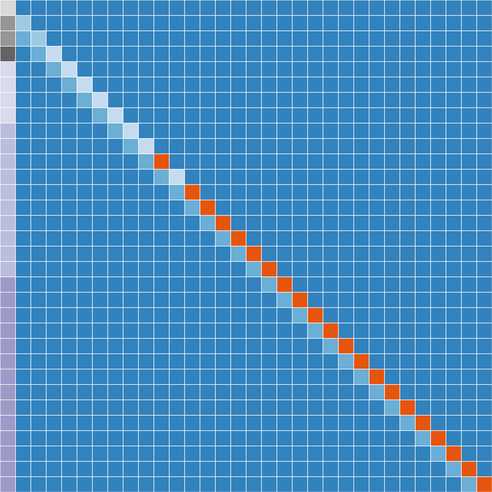}
      & \includegraphics[width=0.15\linewidth]{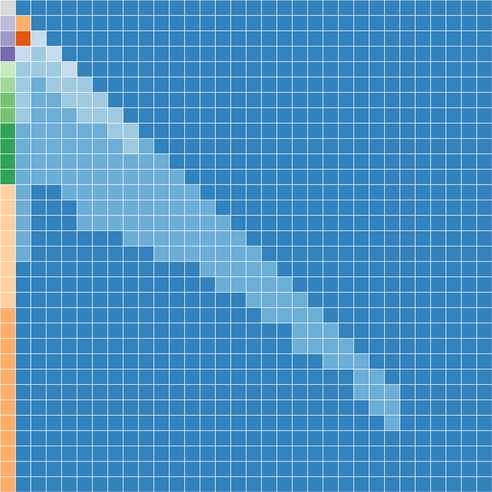} \\
    \hline

    \rotatebox{90}{\textbf{box-cox}}
      & \includegraphics[width=0.15\linewidth]{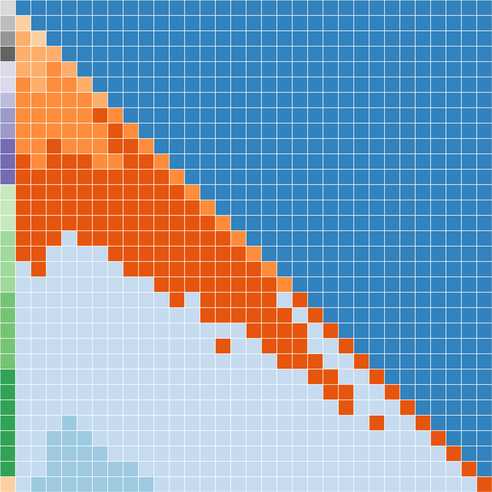}
      & \includegraphics[width=0.15\linewidth]{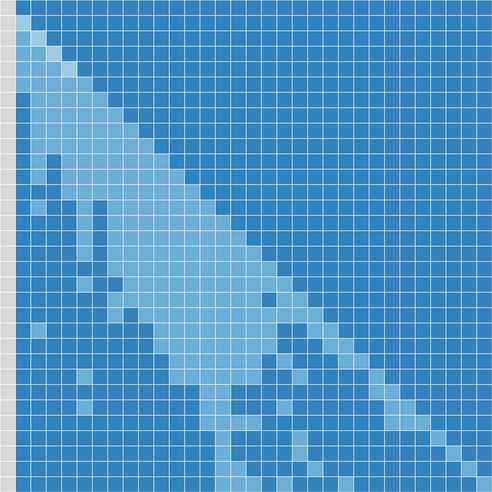}
      & \includegraphics[width=0.15\linewidth]{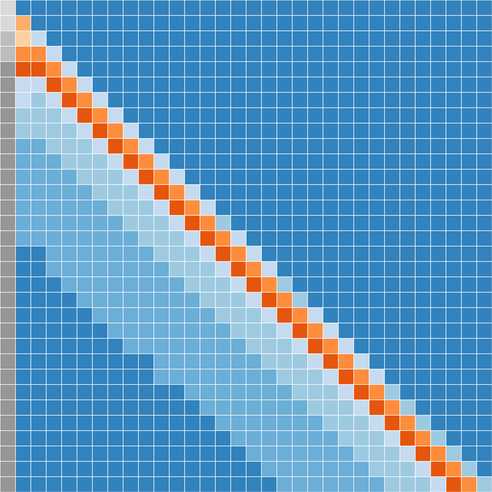}
      & \includegraphics[width=0.15\linewidth]{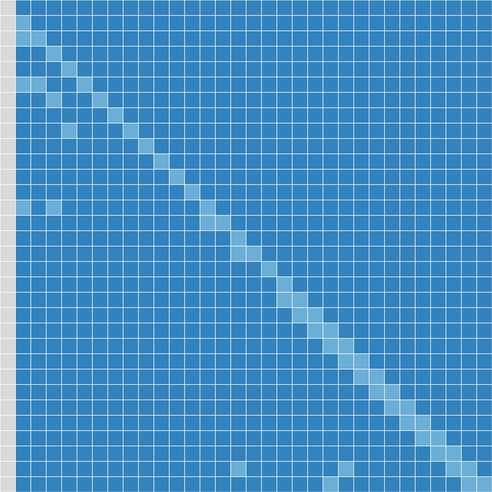}
      & \includegraphics[width=0.15\linewidth]{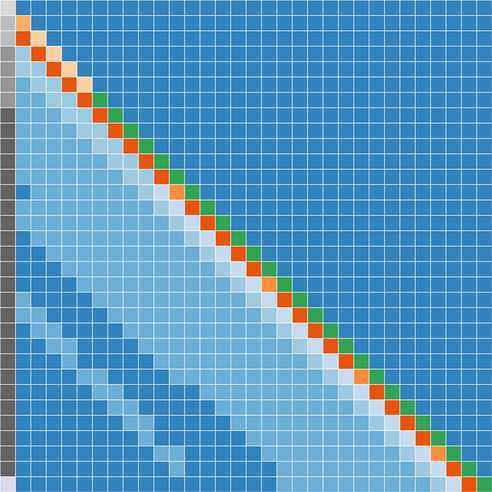}
      & \includegraphics[width=0.15\linewidth]{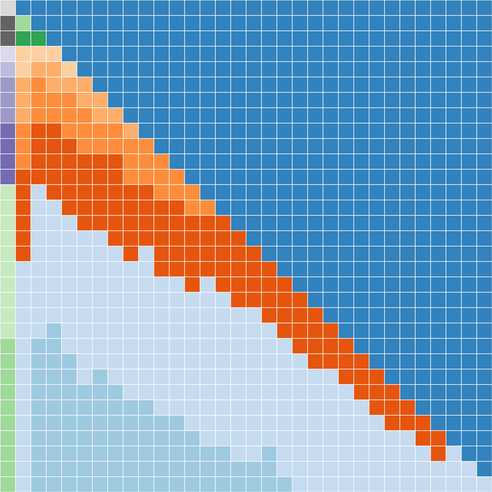} \\
    \hline

    \rotatebox{90}{\textbf{square-root}}
      & \includegraphics[width=0.15\linewidth]{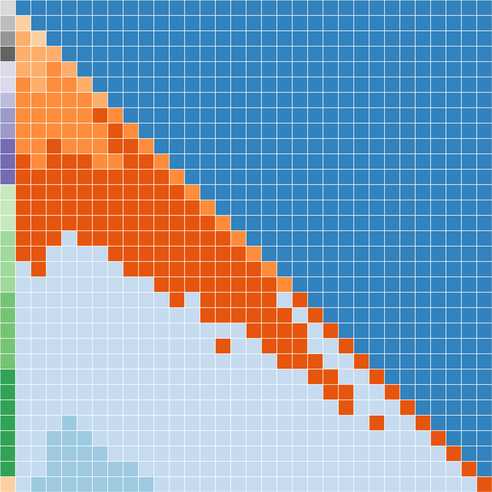}
      & \includegraphics[width=0.15\linewidth]{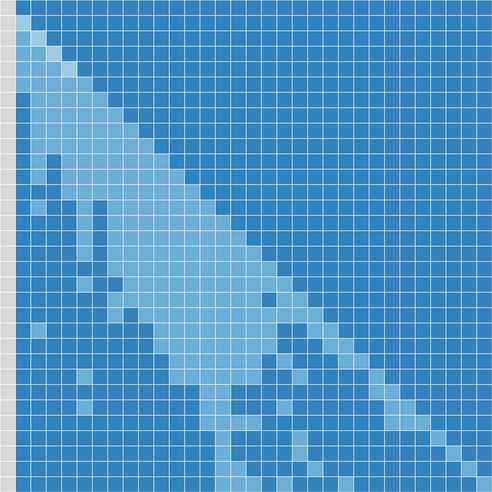}
      & \includegraphics[width=0.15\linewidth]{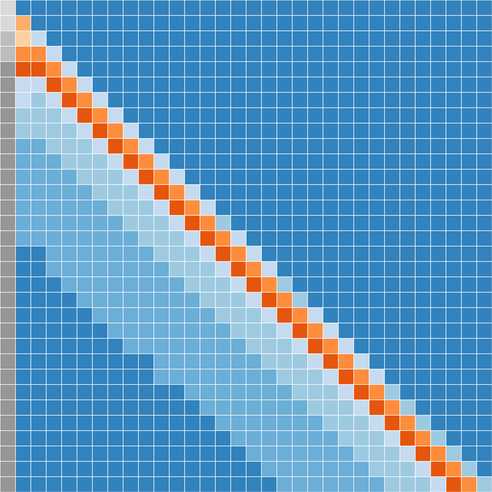}
      & \includegraphics[width=0.15\linewidth]{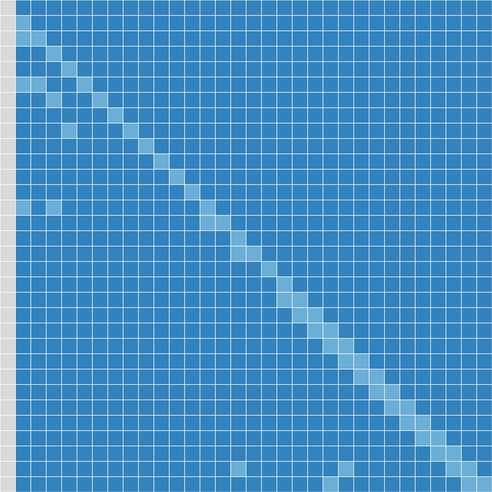}
      & \includegraphics[width=0.15\linewidth]{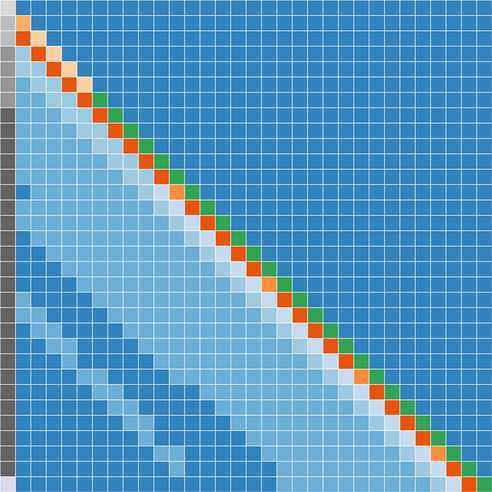}
      & \includegraphics[width=0.15\linewidth]{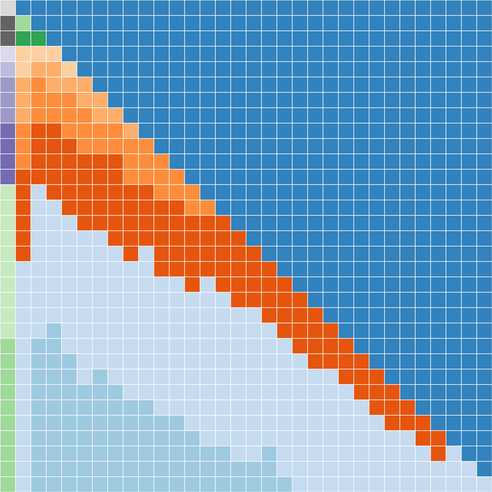} \\
    \hline

    \rotatebox{90}{\textbf{arcsinh}}
      & \includegraphics[width=0.15\linewidth]{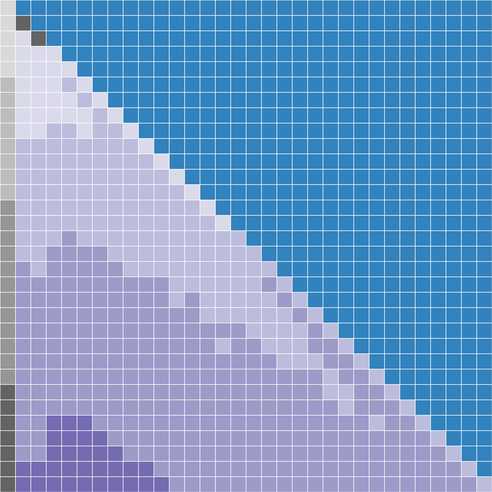}
      & \includegraphics[width=0.15\linewidth]{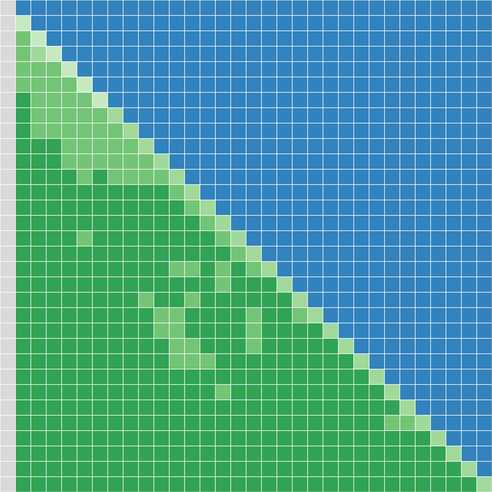}
      & \includegraphics[width=0.15\linewidth]{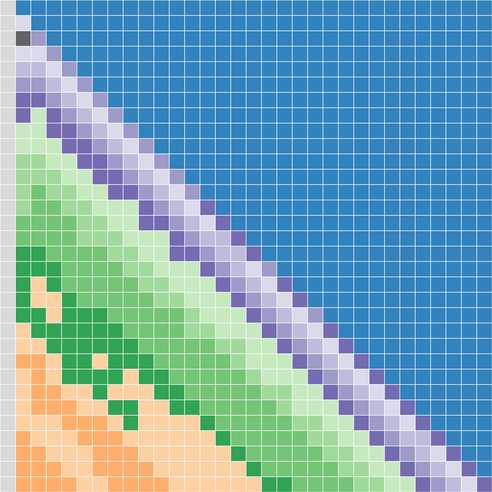}
      & \includegraphics[width=0.15\linewidth]{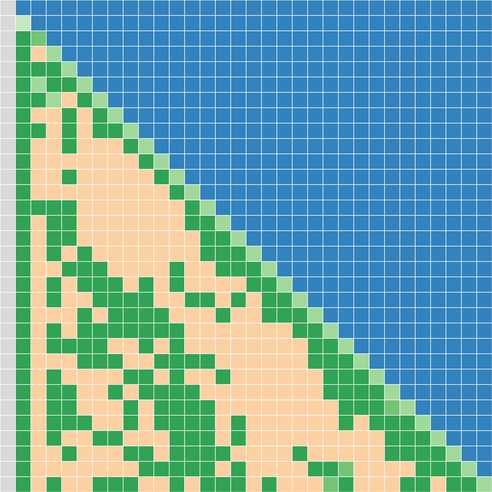}
      & \includegraphics[width=0.15\linewidth]{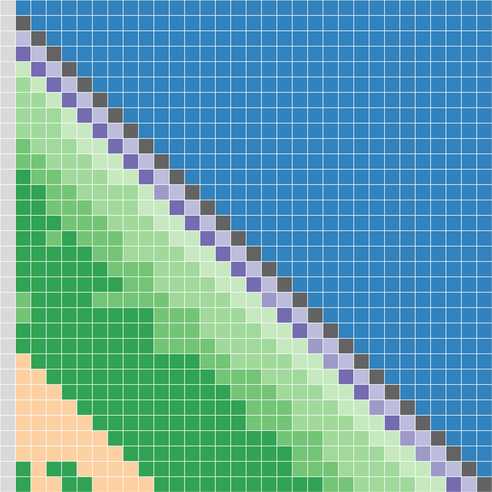}
      & \includegraphics[width=0.15\linewidth]{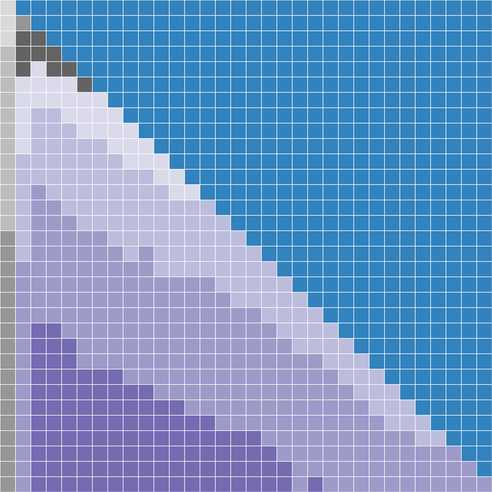} \\
    \hline

    \rotatebox{90}{\textbf{log}}
      & \includegraphics[width=0.15\linewidth]{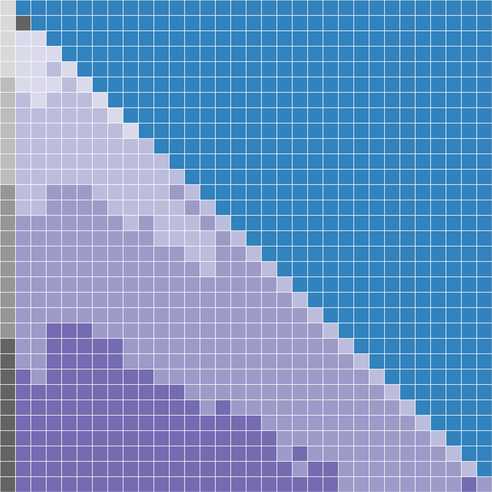}
      & \includegraphics[width=0.15\linewidth]{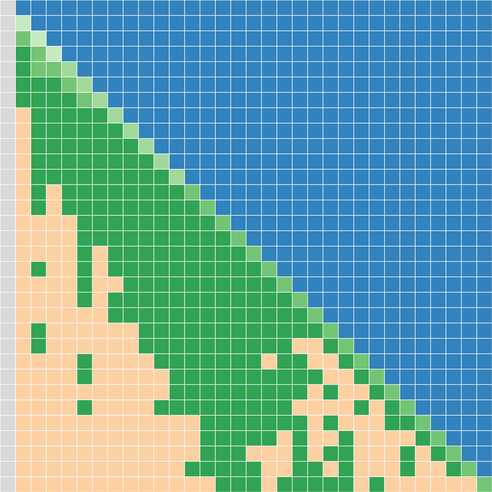}
      & \includegraphics[width=0.15\linewidth]{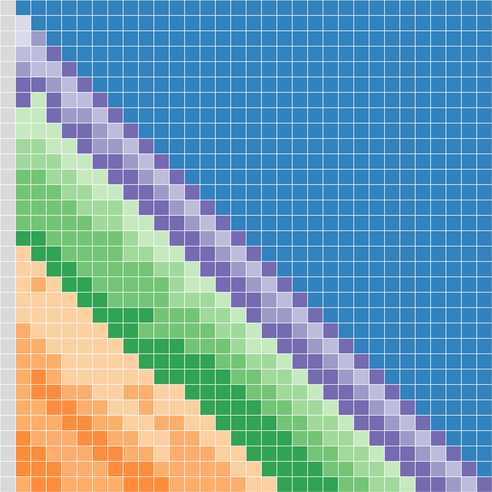}
      & \includegraphics[width=0.15\linewidth]{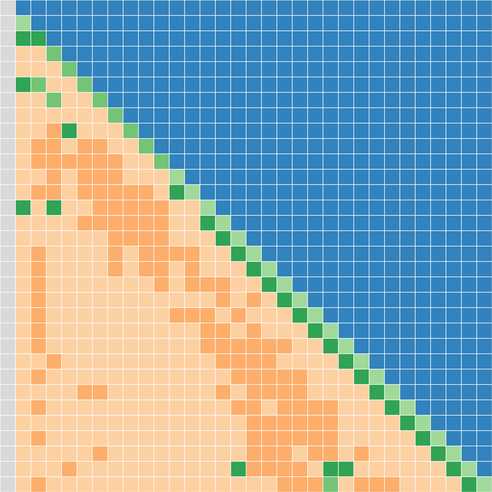}
      & \includegraphics[width=0.15\linewidth]{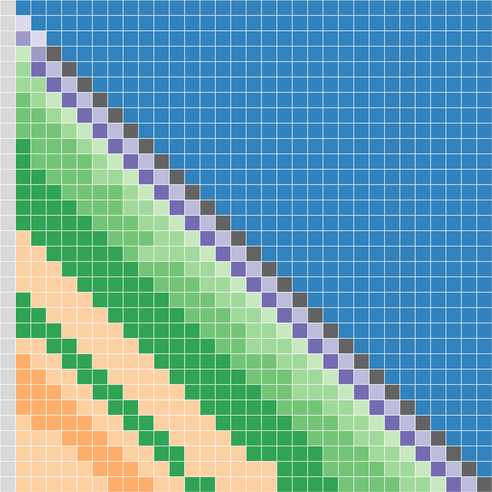}
      & \includegraphics[width=0.15\linewidth]{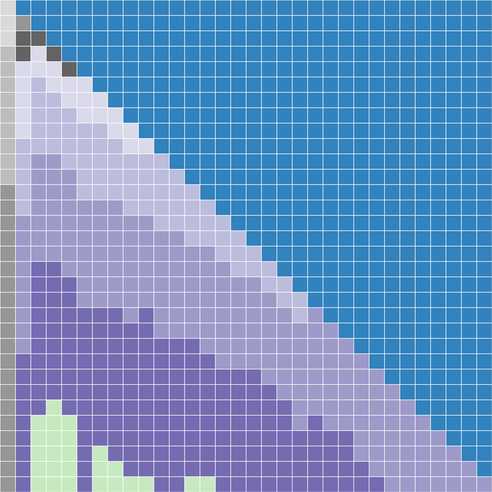} \\
    \hline

  \end{tabular}
  \caption{Feature amplification examples across six attention maps from the LLaMA 3.2 3B model under nine transformation methods. Column headers indicate the layer and head indices of each attention map; row headers correspond to the transformation methods. Box-Cox and Square Root transformations yield more uniform attention value distributions.}
  \label{fig:feature_amplification_methods}
\end{figure*}

\begin{figure*}[ht]
    \centering
    \includegraphics[width=1\linewidth]{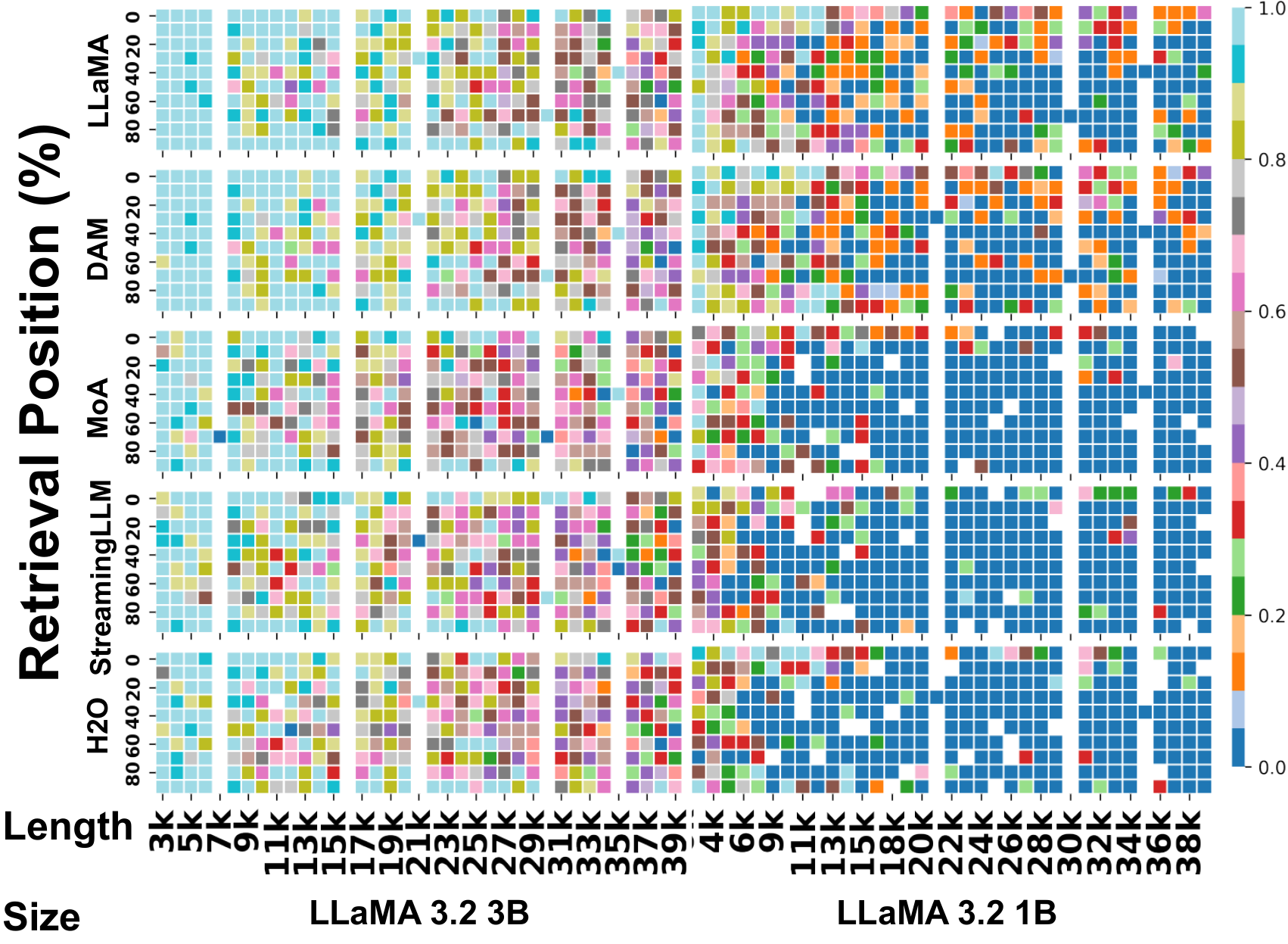}
    \caption{LongEval retrieval accuracy for LLaMA 3.2 3B and 1B models across input lengths to 40K tokens. DAM maintains alignment with the dense LLaMA baseline across retrieval positions and sequence lengths. In contrast, MoA, StreamingLLM, and H2O exhibit early and progressive degradation.}

    \label{fig:LongEval_all}
\end{figure*}

Let $A_{\ell,h,i,j}$ denote the accumulated attention weight from layer $\ell$, head $h$, between token positions $i$ and $j$, and let $C_{\ell,h,i,j}$ be the corresponding valid token count. We compute the average attention as:
\[
\bar{A}_{\ell,h,i,j} = \frac{A_{\ell,h,i,j}}{C_{\ell,h,i,j} + \epsilon}
\]
\noindent where $\epsilon = 10^{-10}$ ensures numerical stability. Let $X = \max(\bar{A}, \epsilon)$ denote the stabilized input. The transformed value is denoted by $\tilde{A}_{\ell,h,i,j}$, and unless otherwise stated, we subtract the global minimum such that $\tilde{A} := \tilde{A} - \min(\tilde{A})$.

\noindent The nine transformation methods are defined as follows:

\noindent \textbf{1. Raw Sum:} $\tilde{A}_{\ell,h,i,j} = A_{\ell,h,i,j}$

\noindent \textbf{2. Average:} $\tilde{A}_{\ell,h,i,j} = \bar{A}_{\ell,h,i,j}$

\noindent \textbf{3. Log:} $\tilde{A}_{\ell,h,i,j} = \log(X)$

\noindent \textbf{4. Box-Cox:} 

\noindent
\quad $\tilde{A}_{\ell,h,i,j} =
\begin{cases}
\frac{X^\lambda - 1}{\lambda}, & \lambda \neq 0 \\
\log(X), & \lambda = 0
\end{cases}$

\noindent \textbf{5. Yeo-Johnson:} 

\noindent 
\quad $\tilde{A}_{\ell,h,i,j} =
\begin{cases}
\frac{(X + 1)^\lambda - 1}{\lambda}, & X \geq 0,\ \lambda \neq 0 \\
\log(X + 1), & X \geq 0,\ \lambda = 0 \\
- \frac{(-X + 1)^{2 - \lambda} - 1}{2 - \lambda}, & X < 0,\ \lambda \neq 2 \\
- \log(-X + 1), & X < 0,\ \lambda = 2
\end{cases}$

\noindent \textbf{6. Z-Score:} $\tilde{A}_{\ell,h,i,j} = \frac{X - \mu}{\sigma + \epsilon}$

\noindent
\quad where $\mu = \text{mean}(X)$, $\sigma = \text{std}(X)$

\noindent \textbf{7. Min-Max:} $\tilde{A}_{\ell,h,i,j} = \frac{X - \min(X)}{\max(X) - \min(X) + \epsilon}$

\noindent \textbf{8. Square Root:} $\tilde{A}_{\ell,h,i,j} = \sqrt{X}$

\noindent \textbf{9. Arcsinh:} 

\noindent
\quad $\tilde{A}_{\ell,h,i,j} = \sinh^{-1}(X) = \log\left(X + \sqrt{X^2 + 1}\right)$

Figure~\ref{fig:feature_amplification_methods} compares the resulting attention maps across six representative heads. The first two rows—\textbf{raw-sum} and \textbf{average}—serve as baselines but fail to reveal an informative structure. Raw-sum maps are dominated by large values in early tokens, while average maps mildly reduce saturation but still obscure subtle patterns, particularly in deeper layers (e.g., L25H9, L27H13).

In contrast, \textbf{box-cox} and \textbf{square-root} transformations enhance interpretability by exposing structural features such as diagonals, stripes, and off-diagonal regions. These patterns are most evident in L3H11 and L15H3, which remain hidden in the baseline maps.

The remaining transformations, including \textbf{yeo-johnson}, \textbf{z-score}, \textbf{min-max}, \textbf{arcsinh}, and \textbf{log}, either overcompress the range or introduce artifacts, resulting in flattened or noisy maps that hinder downstream use.

Table~\ref{tab:transformation_stats} quantifies the numerical differences between square-root and Box-Cox transformations for Layer 25 Head 9. Although both produce visually informative outputs, Box-Cox maps have bounded and compact ranges (e.g., max $\sim$2.0, mean $\sim$0.27), while square-root maps exhibit large variance and extreme values (e.g., max $\sim$150), making thresholding less stable. These results support the use of Box-Cox as the default transformation for attention pattern visualization.

\begin{table}[h]
  \centering
  \begin{tabular}{lcc}
    \toprule
    \textbf{Metric} & \textbf{Square-root} & \textbf{Box-Cox} \\
    \midrule
    Max Value           & 149.95   & 2.00 \\
    Min (non-zero)      & 4.93     & 0.07 \\
    Mean (non-zero)     & 13.57    & 0.27 \\
    Std (non-zero)      & 21.91    & 0.35 \\
    \# Non-zero Values  & 500      & 500 \\
    99th Percentile     & $\sim$100 & $\sim$1.5 \\
    \bottomrule
  \end{tabular}
  \caption{Comparison of square-root and Box-Cox transformed attention values for Layer 25 Head 9. Box-Cox yields compact and stable value ranges that are easier to filter or threshold.}
  \label{tab:transformation_stats}
\end{table}

\section{Long-Context Retrieval}\label{appx:LongEval_all}

We evaluate long-context retrieval using the LongEval benchmark, which measures a model’s ability to recover predefined tokens inserted at various positions within input sequences. Figure~\ref{fig:LongEval_all} presents results up to 40K tokens for LLaMA 3.2 3B and 1B models.

For the 3B models, DAM closely tracks the retrieval accuracy of the dense LLaMA baseline across all lengths. While accuracy gradually declines beyond 30K tokens, DAM preserves similar positional trends. In contrast, MoA, StreamingLLM, and H2O begin diverging much earlier, with noticeable color shifts appearing as early as 3K–7K tokens.

For the 1B models, the differences are more pronounced. LLaMA and DAM maintain high accuracy up to 33K tokens, while MoA and StreamingLLM show early degradation starting around 6K. H2O degrades almost immediately across all target positions.

Overall, DAM preserves fine-grained retrieval performance and retains long-range, position-sensitive dependencies without requiring full attention computation, outperforming alternative efficient methods that degrade under longer contexts.

\end{document}